# Analysis of the Entropy-guided Switching Trimmed Mean Deviation-based Anisotropic Diffusion filter

U. A. Nnolim

Department of Electronic Engineering, University of Nigeria Nsukka, Enugu, Nigeria

## Abstract

This report describes the experimental analysis of a proposed switching filter-anisotropic diffusion hybrid for the filtering of the fixed value (salt and pepper) impulse noise (FVIN). The filter works well at both low and high noise densities though it was specifically designed for high noise density levels. The filter combines the switching mechanism of decision-based filters and the partial differential equation-based formulation to yield a powerful system capable of recovering the image signals at very high noise levels. Experimental results indicate that the filter surpasses other filters, especially at very high noise levels. Additionally, its adaptive nature ensures that the performance is guided by the metrics obtained from the noisy input image. The filter algorithm is of both global and local nature, where the former is chosen to reduce computation time and complexity, while the latter is used for best results.

## 1. Introduction

Filtering is an imperative and unavoidable process, due to the need for sending or retrieving signals, which are usually corrupted by noise intrinsic to the transmission or reception device or medium. Filtering enables the separation of required and unwanted signals, (which are classified as noise). As a result, filtering is the mainstay of both analog and digital signal and image processing with convolution as the engine of linear filtering processes involving deterministic signals [1] [2]. For nonlinear processes involving non-deterministic signals, order statistics are employed to extract meaningful information [2] [3].

Consequently, image noise filtering is a specialized application area of image processing and research work in this field is vast and constantly evolving. Various filter systems are employed for various types of noise, which exhibit certain characteristics. Examples of such noise types include the Additive White Gaussian Noise (AWGN), Multiplicative (speckle) noise, Poisson noise and impulse noise, etc [1] [2] [3]. The best estimators for the Gaussian noise situation are the averaging or mean filters while the median filters are the ideal estimators for the salt and pepper impulse noise [3]. The rule of filter design is to devise a filter such that its noise suppression ability should not be at the expense of signal (edge) preservation capability. However, this is a contradiction for linear smoothing filters, which have low-pass characteristics [2] [3]. In other words, these filters perform indiscriminate smoothing on both signal and noise. This issue becomes a greater concern with increase in noise levels, where much higher degrees of smoothing are required. This is so due to the eventual over-smoothing of edges in the attempts to suppress or eliminate the noise in the image signal. As noise levels increase, the signal becomes highly degraded with subsequent linear filtering, leading to heavy blurring and/or smudging. Though methods such as using Wiener, Bilateral filtering and Fourier or Wavelet domain-based approaches have been utilized for noise types such as the Gaussian type, they do not work for impulse noise, which has no frequency response [1] for example. Furthermore, the Anisotropic Diffusion algorithm has been successfully applied to the filtering of Gaussian [4] and Speckle noise [5]. However, we restrict the focus of the work to salt and pepper noise due to the nonlinear nature of the problem and difficulties encountered with high levels of noise.

## 1.1    Survey of Median-based filters

Median filter and its numerous variants have been shown to be effective at filtering salt and pepper impulse noise. These include the Adaptive Median Filters (AMF) [6] [7] [8], Recursive Median Filters (RMF) [9]- [10], Weighted (stack) Median Filters (WMF) [6], [11], Center Weighted Median Filters (CWMF) [12], Progressive Switching Median Filters (PSMF) [13], Noise Adaptive Fuzzy/Soft-Switching Median Filters (NAFSMF/NASSMF) [14] [15], Fuzzy (Weighted) Median Filters (FMF/FWMF) [16], Median Filtering utilizing Regularization method [17] and Fuzzy FIRE filters [16], [18] to name a few. It is also important to mention the Alpha-trimmed Mean Filter [19]- [20], which is a hybrid filter for filtering impulse noise and additive Gaussian noise depending on the value of alpha parameter.

Some Fuzzy variants utilizing conventional switching schemes [21] [22] [23] [24] [25] [26][18]-[23] use a hard or soft limiter mechanism based on threshold matching the Median Absolute Deviation parameter to gauge the filtering process. However, these work best at low to medium noise density levels and successive application leads to excessive smudging of edges, causing signal degradation. Decision-based and Fuzzy-logic-based filters were developed to address this issue of increased noise density applications [13] [16]. However, these also have a limit in the medium noise density levels, beyond which the fuzzy-based filters fail. Other works have increased the complexity of such median-based filters to address higher density noise situations.

However, these also fail at medium to relatively high levels of noise. Thus, works by [17] and others using Fuzzy Cellular Automata (FCA) [27] have been reported in the literature for improving results. Additionally, trimmed mean filters have also been used for high density impulse noise filtering [28]. As such, there are numerous median-based filters and variants with varying levels of computational and structural complexity and mixed results that it becomes difficult to fully compare all the recent developed algorithms in the literature. However, based on experiments, it can be said that the median filter has reached its limit. Thus, a new approach is required to push the performance of filters for impulse noise filtering.

## 1.2    Brief overview of Anisotropic Diffusion

Anisotropic diffusion basically involves the modification of the isotropic (equal energy in all directions) heat equation given as shown in (1) [4] [5];

$$\frac{\partial U(x,y,t)}{\partial t} = D\nabla U(x,y) \quad (1)$$

into an anisotropic one where energy is minimized by only allowing diffusion along the edges, resulting in signal preservation [4] [5], yielding the expression in (2);

$$\frac{\partial U(x,y,t)}{\partial t} = \nabla.\left(D(s)\nabla U(x,y)\right) \quad (2)$$

In both equations, $D$ is the diffusion coefficient and $\nabla U(x,y)$ is the image gradient. However, the diffusion coefficient is a constant in (1) while it is a varying function of $s$ in (2). The original functions used by Perona and Malik for $D$ are the Gaussian and Cauchy functions [4] respectively as shown in (3) and (4).

$$D(s) = e^{-\left(\frac{s}{\kappa}\right)^2} \quad (3)$$

$$D(s) = \frac{1}{1+\left(\frac{s}{\kappa}\right)^2} \quad (4)$$

In (3) and (4), $s = |\nabla U|$ while $\kappa$ is the diffusion threshold parameter [4] [5]. There have been several modifications of the standard approach over the years [29]. One such method proposed a modified diffusion coefficient function obtained by a regularized AD function, in the form [30] shown in (5);

$$s = |\nabla G_\sigma * U| \quad (5)$$

In (5), $G_\sigma$ is a Gaussian smoothing kernel, with width or standard deviation, $\sigma$ and $*$ indicates convolution. The modified term is reported to yield a better estimate for local gradient for Gaussian noise [29]. However, the blurring effect, which is a main feature of the regularization operation, is observed. Alternative ways of expressing AD involve using the div operator [29] as shown in (6);

$$\frac{\partial U(x,y,t)}{\partial t} = div(D(|\nabla U|)\nabla U) = \nabla D . \nabla U + D\Delta U \quad (6)$$

The regularization version as described in [29] is given as;

$$\frac{\partial U(x,y,t)}{\partial t} = div(D(|\nabla U_\sigma|)\nabla U) = \nabla D . \nabla U_\sigma + D\Delta U \quad (7)$$

Where $U_\sigma(x,y,t) = G_\sigma * U(x,y,t)\sigma$ and the diffusion function [29]- [30] is given as

$$D(s) = \begin{cases} \left[1 - \left(\frac{s}{\kappa}\right)^2\right]^2 & if \ |s| \leq \kappa \\ \\ 0 & otherwise \end{cases} \quad (8)$$

Another improved AD formulation [29] is given as;

$$\frac{\partial U(x,y,t)}{\partial t} = div\left((D(|\nabla U|) - \upsilon(|\nabla U|))\nabla U\right) \quad (9)$$

Where $D(|\nabla U|)$ is the diffusion coefficient and $\upsilon(|\nabla U|)$ is the sharpen coefficient [29], enabling the expression in (9) to perform simultaneous sharpening and smoothing [29]. The next improved P-M AD model proposed by Sum and Cheung [29] [31] is given as;

$$\frac{\partial U(x,y,t)}{\partial t} = div(D(|\nabla U|)\nabla U) + \frac{1}{2}|U - U_o| \quad (10)$$

However, as noted in [29], the formulation in (10) does not guarantee robust restoration solution and image edges are not enhanced. Thus, the work of Zhou and Liu [32] goes further by using not only $U_\sigma(x, y, t) = G * U(x, y, t)$ for the regularization operation but in addition to minimization of energy functional in image domain, $\Omega$ [29] [32] given as;

$$E(U) = \int_\Omega \rho(|\nabla U_\sigma|) \, dx \, dy + \lambda \int_\Omega |U - U_o| \, dx \, dy \quad (11)$$

Where $\rho(|\nabla U_\sigma|) = D(|\nabla U_\sigma|)|\nabla U_\sigma|$ is the gradient drop flow and $U_o = U(x, y, 0)$, ultimately leading to the gradient flow equation [29] given as;

$$\frac{\partial U(x,y,t)}{\partial t} = \text{div}(D(|\nabla U|)\nabla U) - \lambda(U - U_o) \quad (12)$$

Additionally, Rudin and Osher also proposed the Shock Filter PDE-based model for image enhancement [33][28], which is concerned with sharpening and smoothing. Another popular and well-established PDE-based model is the total variation approach with regularization parameter proposed by Rudin, et al [34]- [35]. However, the standard P-M model yields better edge preservation than the standard TVR method, though there have been improvements based on the works of several authors [29].

With these numerous variations of the original PM AD model, it is important to note that all these are suited to filtering images corrupted with Gaussian noise, though there are few relatively recent works utilizing PDE-based methods for impulse noise filtering [17], [36]- [37]. Additionally, for this work, we slightly modify the equations in order to use the PDE for salt and pepper impulse noise. The proposed algorithm is essentially of two parts, which is common for the PDE-based approaches; namely an impulse noise detector section and a pixel replacement or filtering section. The subsequent sections will discuss both parts in detail.

## 2. Proposed impulse noise detection and modification filter

The We now describe the nature of the first component of the proposed algorithm, which is the Switching Trimmed Mean Deviation Filter (STMDF).

### 2.1 Switching Trimmed Mean Deviation Filter (STMDF)

Given an ordered set, $X = \{x_1, x_2, \ldots, x_j\}$, the trimmed mean, $\widehat{x_m}$ is given as;

$$\widehat{x_m} = \frac{1}{n - 2[\propto n] + 1} \sum_{j=[\propto n]+1}^{n-[\propto n]+1} x_j \quad (13)$$

Consequently, the sample trimmed median deviation vector is given as;

$$\delta_{STMD} = \{\widehat{x_m} - x_j\} \quad (14)$$

Additionally, the trimmed mean absolute deviation yields;

$$TMAD = |\widehat{x_m} - X| \quad (15)$$

and the centered trimmed mean deviation yields;

$$\delta_{CTMD} = \widehat{x_m} - x_c \qquad (16)$$

Combined with the switching scheme yields;

$$y_c = \begin{cases} x_c; & |\delta_{CTMD}| \leq \tau \\ \widehat{x_m}; & otherwise \end{cases} \qquad (17)$$

Where the parameter, $\tau = (\mu_g - \sigma_g) * entropy$ is the entropy guided threshold, while the entropy $= \sum_i^B (-p_i * \log(p_i))$, where $p_i$ is the normalized histogram of the image of bin element, $i$ while $B$ is the total number of bins. Additionally, $\mu_g = \frac{1}{MN} \sum_{m=1}^{M} \sum_{n=1}^{N} I(m,n)$ is the global mean while $\sigma_g = \sqrt{\frac{1}{MN} \sum_{m=1}^{M} \sum_{n=1}^{N} [I(m,n) - \mu_g]^2}$ is the global standard deviation from the global mean. The local mean and standard deviation could also be used in this expression, though it would considerably increase computation time. The next step is to combine this formulation with the Anisotropic Diffusion filter to guide the filtering along the edges.

## 2.2    Entropy-based STMDF-Anisotropic Diffusion filter

Following from the work discussed in the introduction, we proceed from the formulation proposed by [5] expressed in the form;

$$\frac{\partial U(x,y,t)}{\partial t} = \nabla.(D\nabla U) + \alpha f(x,y,t) \qquad (28)$$

The discretization of the expression leads to the form shown in (29);

$$U^{t+1} = U^t + (\nabla.(D(s)\nabla U) + \alpha f(x,y,t))\Delta t \qquad (29)$$

In (28) and (29), $f(x,y,t) = \text{MedFilt}(U(x,y,t))$, while $D(s)$ is the Cauchy function in (3) and for the proposed approach, we also ensure that the diffusion threshold, $\kappa$ is computed as a function of image statistical parameters in the form;

$$\kappa = \frac{\mu}{\sigma} \qquad (31)$$

Multiplying through by $\Delta t$ and setting $\Delta t = 1/4$ and $\beta = \alpha/4$ as shown in [23] we obtain the expression in (30), which is the proposed approach where $f(x,y,t) = \text{STMDF}(U(x,y,t))$.

$$U^{t+1} = (1-\beta)U^t + [\nabla.(D(s)\nabla U)]/4 + \beta f(x,y,t) \qquad (30)$$

This enables the computation of the continuous image.

## 2.3 Total Variation Regularized (TVR) STMDF

Another alternative was using the TVR approach similar to [34].

$$\frac{\partial U(x,y,t)}{\partial t} = \nabla \left( \frac{\nabla U}{|\nabla U_\varepsilon|} \right) + \lambda(U_o - U) \qquad (31)$$

Where $U_o = U(x, y, 0) = U_o(x, y)$, which is the initial noisy image and $|\nabla U_\varepsilon| = \sqrt{U_x^2 + U_y^2 + \varepsilon^2}$ is the norm where

$$U_x = \frac{\partial U(x,y)}{\partial x}; \ U_y = \frac{\partial U(x,y)}{\partial y} \qquad (32)$$

Leading to the discrete form;

$$U^{t+1} = U^t + \left[\nabla\left(\frac{\nabla U}{|\nabla U_\varepsilon|}\right) + \lambda(U_o - U)\right]\Delta t \qquad (33)$$

With

$$\nabla\left(\frac{\nabla U}{|\nabla U_\varepsilon|}\right) = \frac{(U_x^2 + \varepsilon^2)U_{yy} + (U_y^2 + \varepsilon^2)U_{xx} - 2U_x U_y U_{xy}}{\left(U_x^2 + U_y^2 + \varepsilon^2\right)^{3/2}} \qquad (34)$$

With $U_{xx} = \frac{\partial^2 U(x,y)}{\partial x^2}$; $U_{yy} = \frac{\partial^2 U(x,y)}{\partial y^2}$; $U_{yy} = \frac{\partial^2 U(x,y)}{\partial x \partial y}$ and $\varepsilon$ is a very small number to prevent division by zero. The initial formulation works well for additive and speckle (multiplicative noise). Thus for impulse noise, we reformulate the expression as;

$$\frac{\partial U(x,y,t)}{\partial t} = \nabla\left(\frac{\nabla U}{|\nabla U_\varepsilon|}\right) + \lambda(f - U) + \alpha f \qquad (35)$$

Leading to the discrete form;

$$U^{t+1} = U^t + \left[\nabla\left(\frac{\nabla U}{|\nabla U_\varepsilon|}\right) + \lambda(f - U)\right]\Delta t + \alpha f \qquad (36)$$

As in (28) - (30), expression in (35) - (36) the term, $f(x, y) = STMDF\big(U(x, y)\big)$. However, this method does not perform well and needs a kernel of size 7×7 for reasonable results, based on extensive experiments. The approach was abandoned due to sheer computational effort with marginal results. The Anisotropic Diffusion appears to be better suited to the problem than the TVR formulation. So far all AD-based methods have surpassed TVR-based median filter approaches. Median-based STMDF preserves edges better reducing global blurring but yields lower PSNR, especially at higher noise densities while the trimmed mean-based STMDF blurs images, but yields higher PSNR, especially at higher noise densities and coupled with the Gaussian edge-preserving function rather than the Cauchy function.

The choice of using default 3×3 spatial neighbourhoods for both STMDF and AD makes the design stable. Increasing the window size leads to considerable blurring, loss of edge information and details with high degree of smudging. The following section deals with experiments to verify the theoretical formulations and justifications used to develop the proposed approach.

## 3 Experiments and results

This section presents the image data, the prior experiments aimed at estimation of image statistical parameters and their relationships with regards to noise to enable a more accurate function for noise estimation and detection. Fig. 1 shows the sample images used for experiments while Fig. 2 shows the various plots relating the image parameters to noise density.

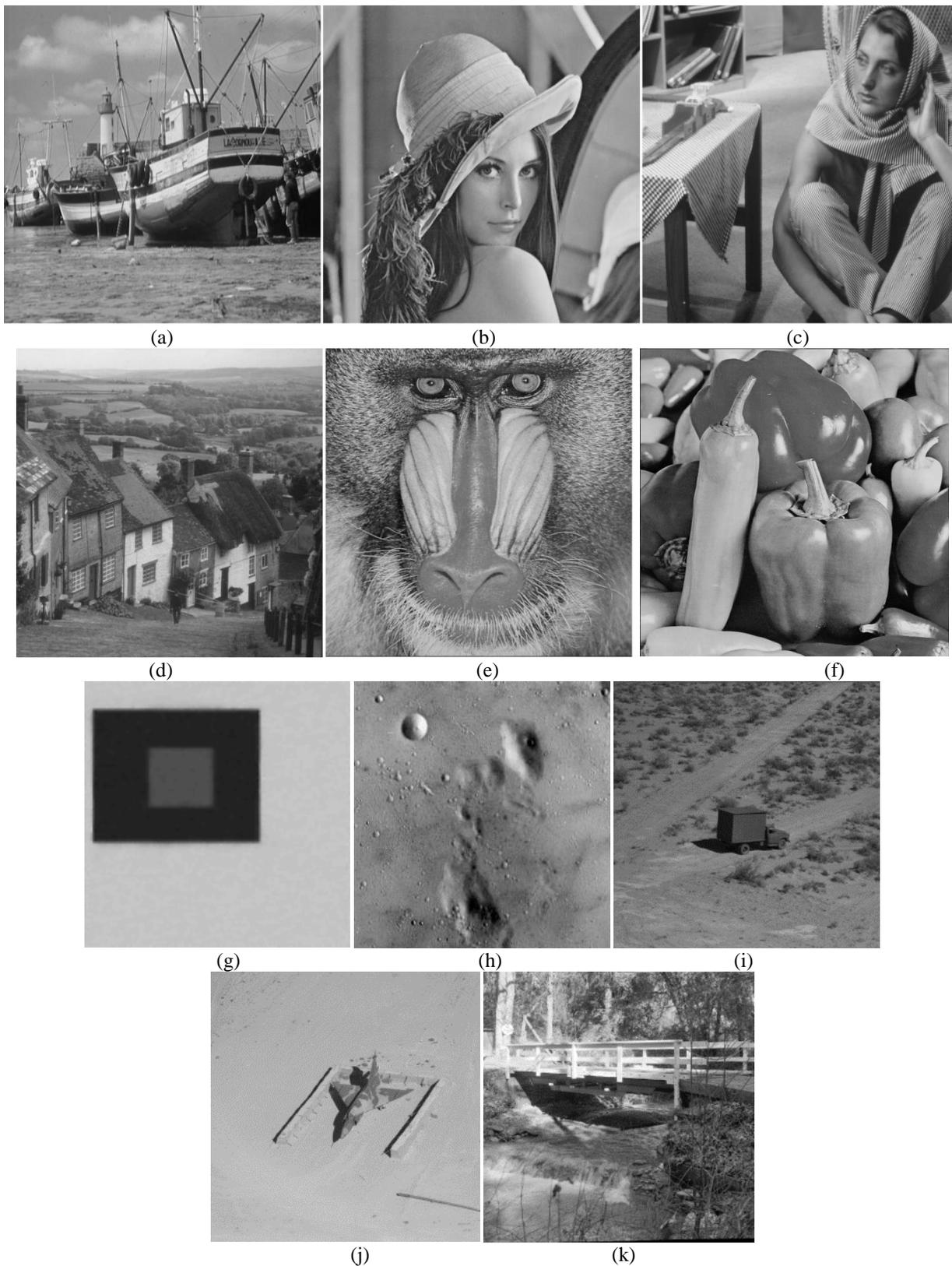

**Fig. 1** 8-bit sample greyscale images used for experiments: (a) Boat (b) Lena (c) Barbara (d) Gold Hill (e) Baboon (f) Peppers (g) Chip (128×128) (h) Moon (256×256) (i) Truck (k) Jet (f) Bridge (512×512)

The problem is to determine the best function that relates the standard deviation, threshold, entropy to yield a more accurate noise density estimate. Numerous images were tested and the graphs obtained were mostly consistent. In the plots, the most reliable predictors of noise density trends tend to be the standard deviation and entropy parameters. The mean appears to be the least reliable predictor/estimator of noise density, and appears erratic, unstable and unreliable as a noise estimation parameter.

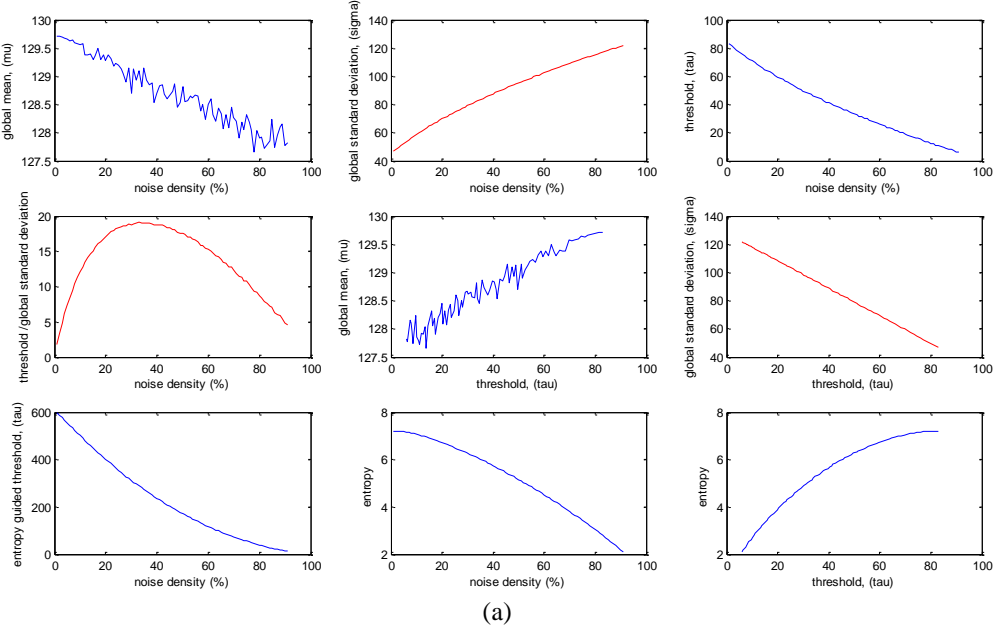

(a)

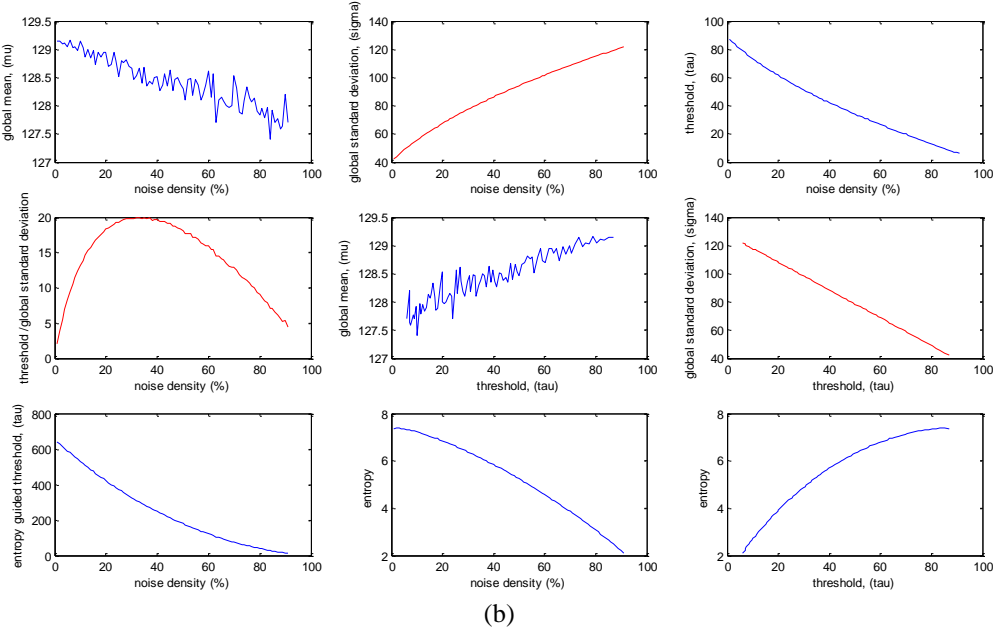

(b)

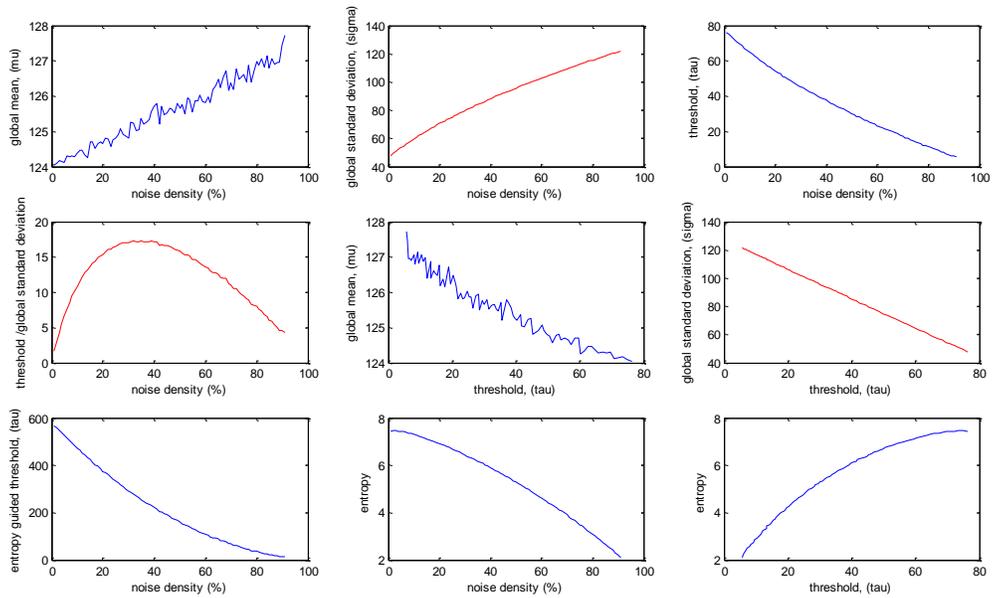

(c)

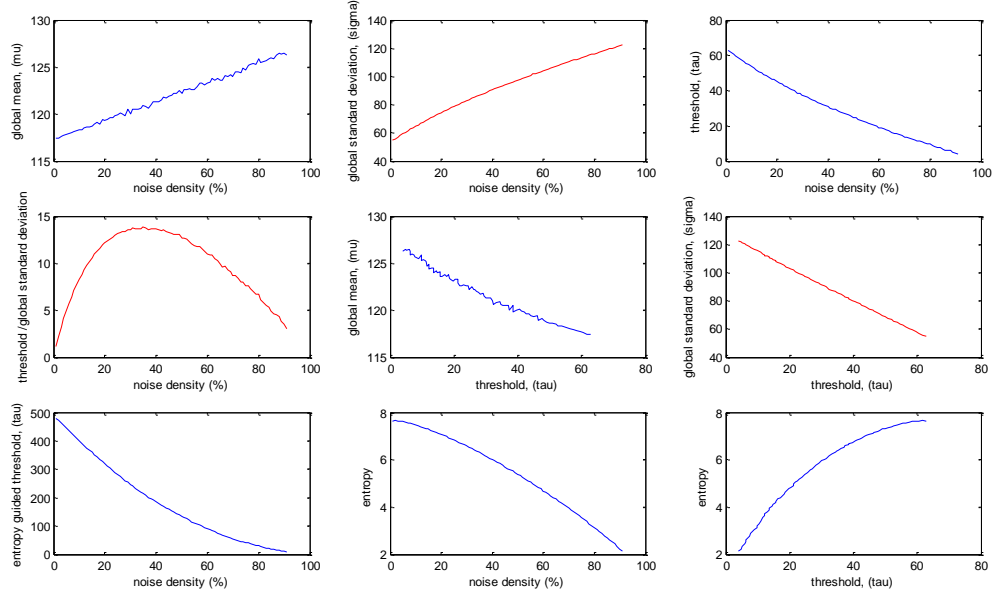

(d)

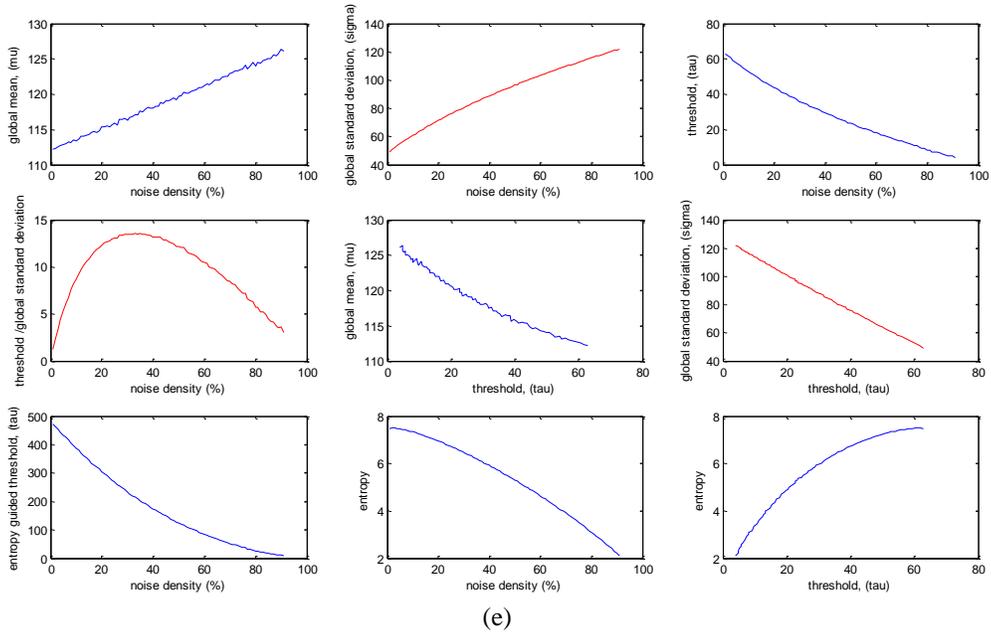

(e)

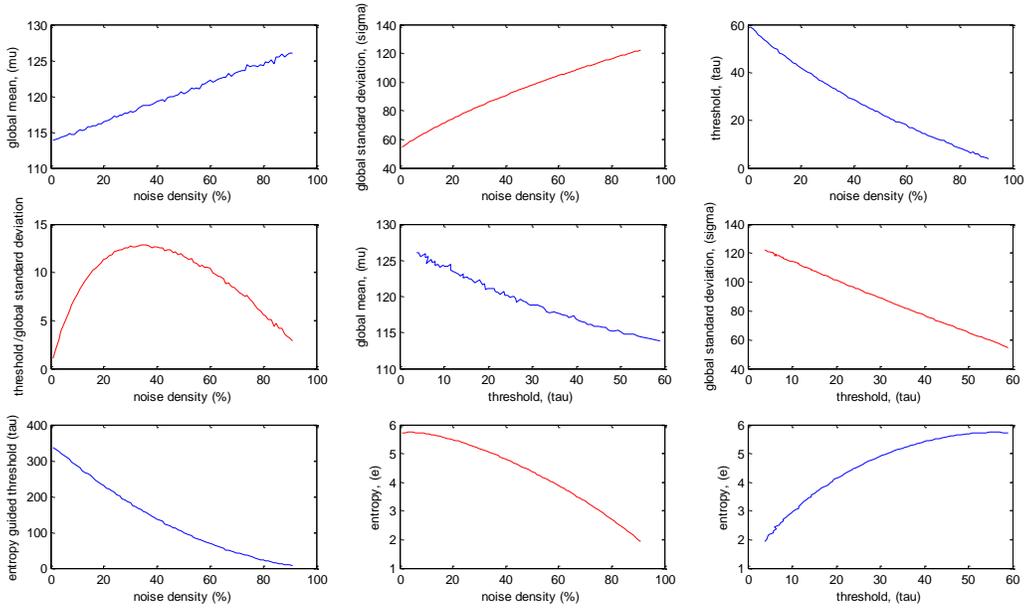

(f)

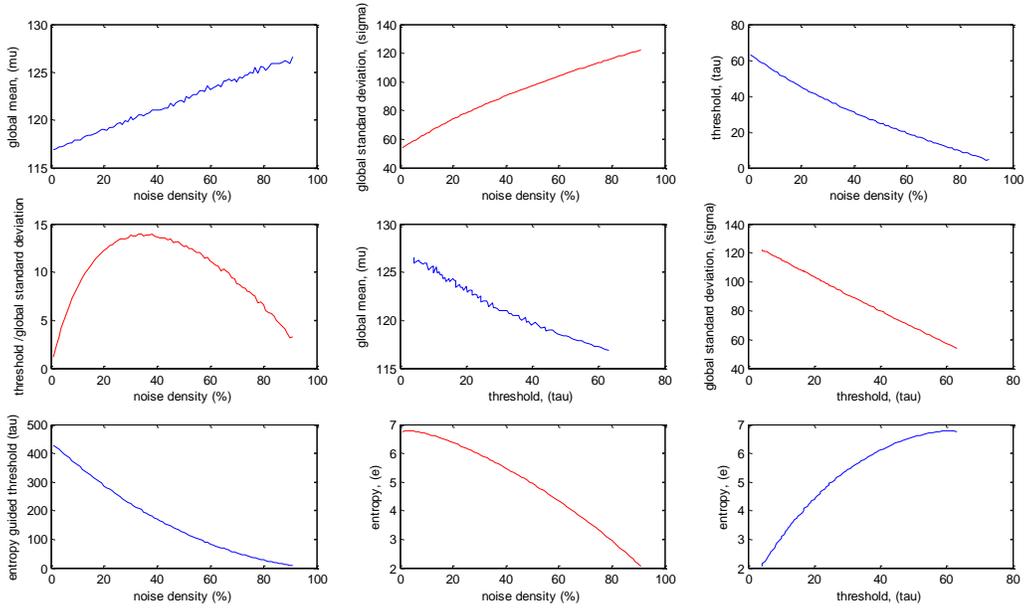

(g)

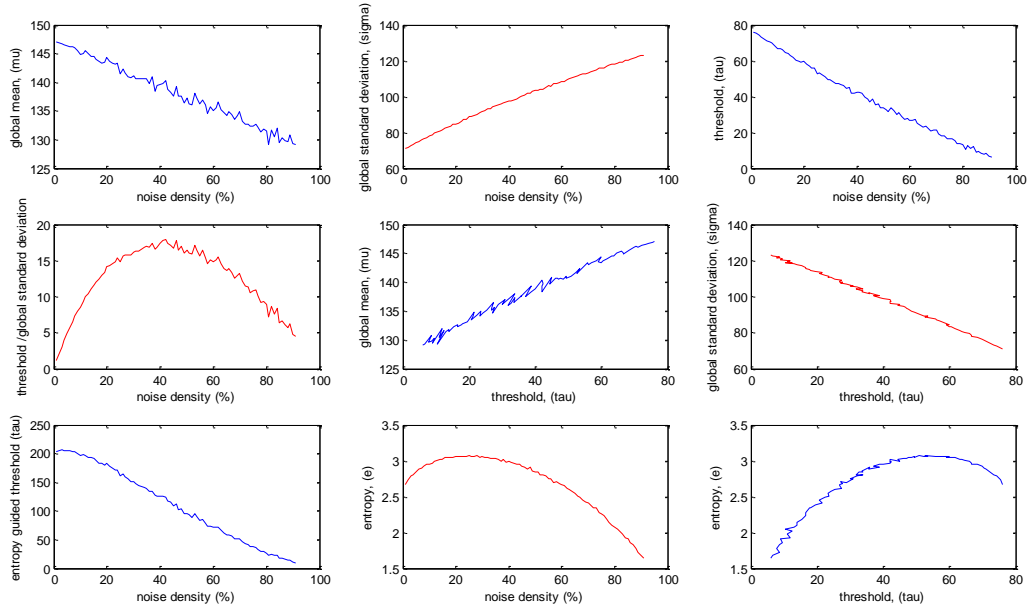

(h)

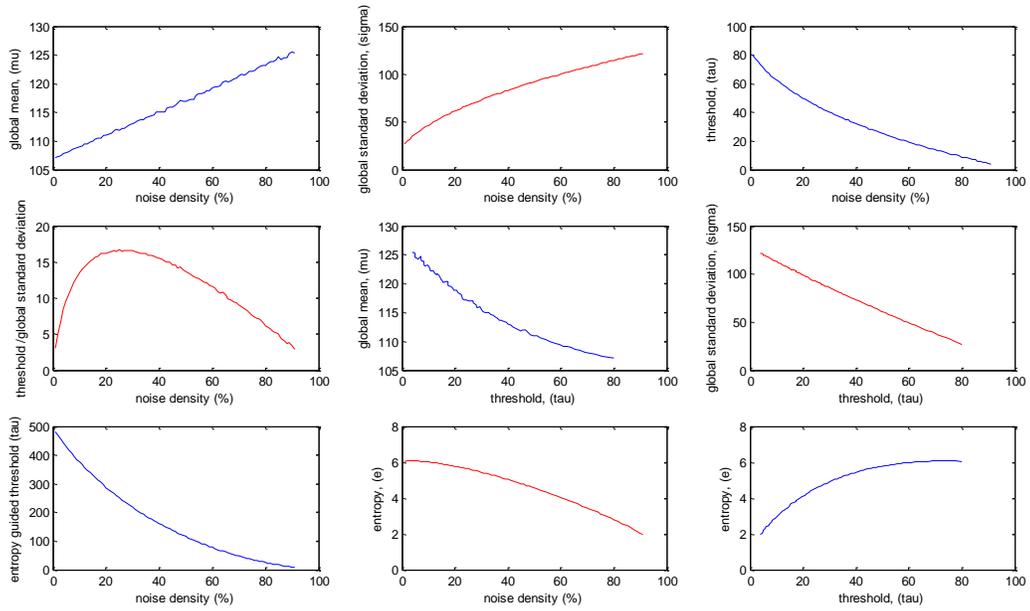

(i)

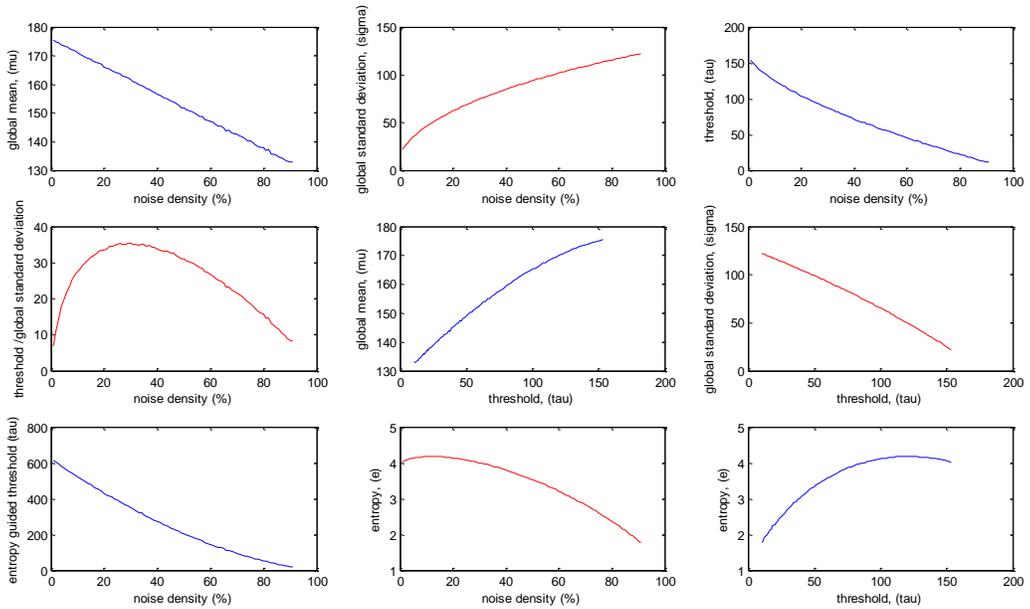

(j)

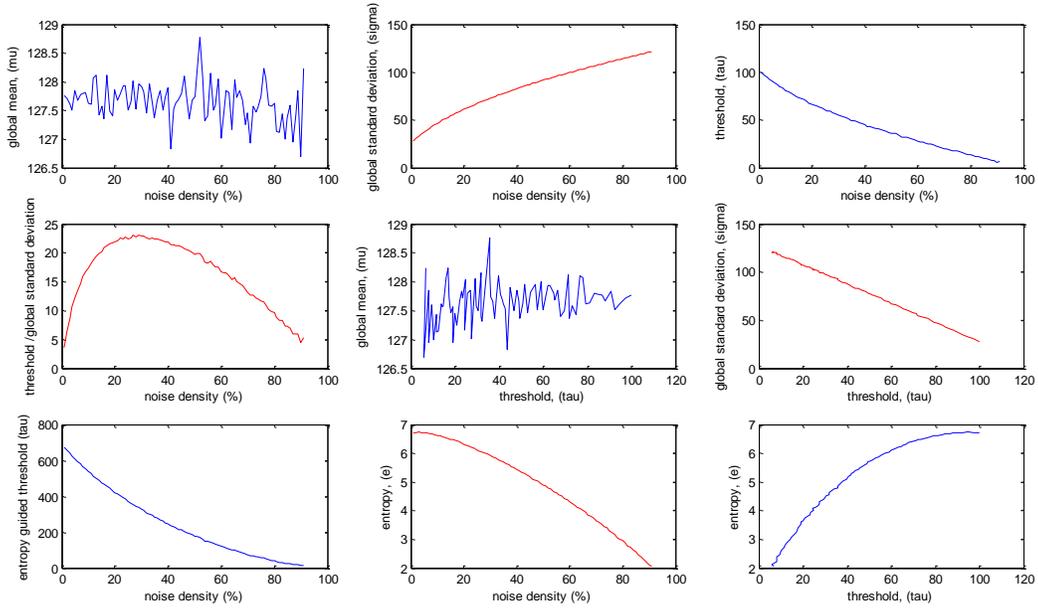

(k)

**Fig. 2** plots of image attributes for (a) Boat (Entropy = 7.1914) (b) Baboon (Entropy = 7.3579) (c) Lena (Entropy = 7.4455) (d) Barbara (Entropy = 7.6321) (e) Gold Hill (Entropy = 7.4778) (f) Bridge (Entropy = 5.7056) (g) Peppers (Entropy = 6.7624) (h) Chip (Entropy = 2.6725) (i) Truck (Entropy = 6.0274) (j) Jet (Entropy = 4.0045) (k) Moon (Entropy = 6.7093) images

Initially, experiments were performed with the median filter-boosted anisotropic diffusion. However, the results were unacceptable, especially at high noise densities. This can be seen in Fig.3 and Table 1 showing that a great deal of information is lost and thus, the MF-AD cannot recover much of the original image at high noise densities. At a lower noise density of 50%, five iterations yield some image restoration, which is quite poor compared to results obtained from other median filter variants from the literature. We also utilize a NASMF-boosted AD to compare results and this formulation still did not yield much improvement at higher densities.

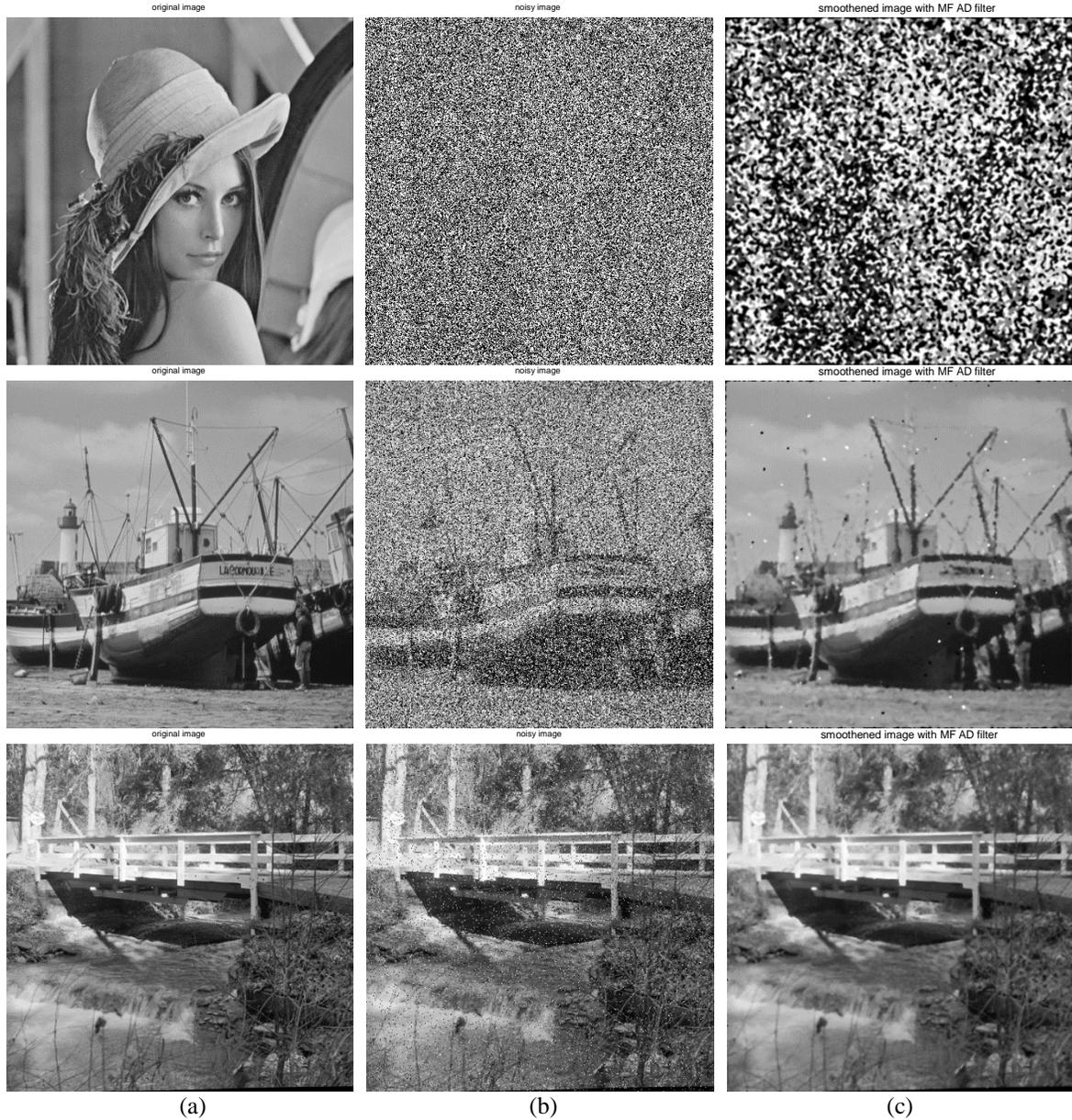

|  | (a) | (b) | (c) |

**Fig. 3** (a) Original Lena (first row), Boat (second row) and Bridge images (third row) (b) corrupted by 90% (first row), 50% (second row) and 5% (third row) salt and pepper FVIN (c) filtered with MF-AD

| Measures | Lena (90% FVIN) | Boat (50% FVIN) | Bridge (5% FVIN) |
|----------|-----------------|-----------------|------------------|
| PSNR (dB) | 9.5133 | 24.3647 | 25.4816 |
| MAE | 67.0099 | 7.6779 | 9.0813 |
| MSE | 7.2736e+003 | 238.0204 | 184.0437 |
| MSSIM | 0.04981 | 0.75954 | 0.72310 |

**Table 1**. Image quality metrics for processing corrupted Lena, Boat and Bridge images with median boosted Anisotropic Diffusion filter

Finally, a combined STMDF-based Total Variation Regularized (TVR-STMDF) variant was implemented to compare with the AD-based approaches. The results in Fig. 4 show that the TVR-STMDF yields the worst results at high noise densities in addition to a large number of iterations and further increased computational complexity. Thus, this scheme was abandoned and the STMDF-AD was selected due to its consistency of quality results and relative speed, requiring fewer iterations.

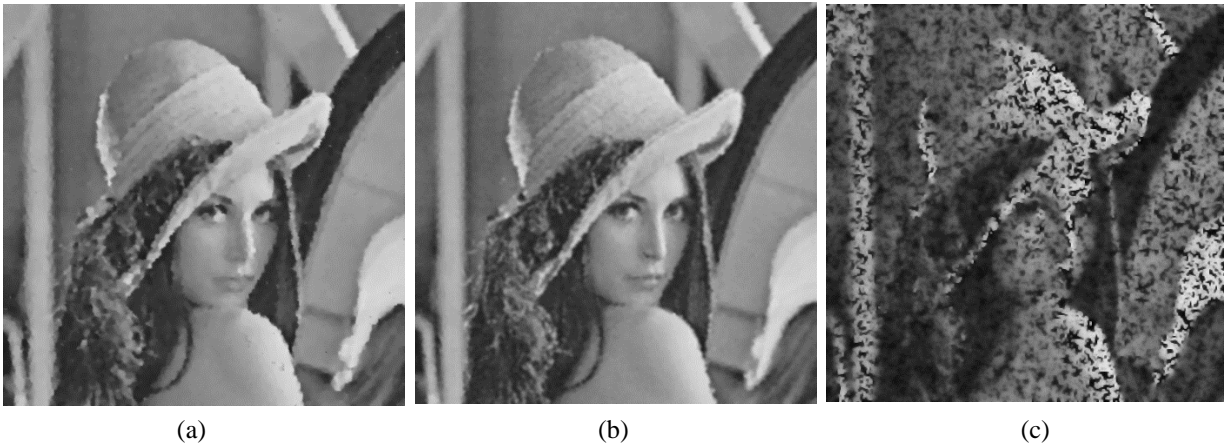

(a)                                      (b)                                      (c)

**Fig. 4** Filtered Lena image (corrupted with 90% salt and pepper FVIN) using (a) NAFSMF-AD and (b) STMDF-AD (c) TVR-STMDF

| Measures | TVR-STMDF | MF-AD | NAFSMF-AD | STMDF-AD |
|----------|-----------|-------|-----------|----------|
| PSNR (dB) | 5.5771 | 9.5133 | 26.2880 | **27.3223** |
| MAE | 116.6913 | 67.0099 | 6.6488 | **6.0540** |
| MSE | 1.8004e+04 | 7.2736e+003 | 152.8552 | **120.4627** |
| MSSIM | 0.00856 | 0.04981 | 0.76621 | **0.79663** |

**Table 2**. Quantitative metrics for Filtered Lena image (corrupted with 90% salt and pepper FVIN) using MF-AD, NAFSMF-AD and STMDF-AD

## 3.1 Filter performance comparisons

Fig. 3 to 4 and Table 1 to 2 show that the STMDF-AD formulation yields much better results than the other AD combinations. This section compares the results of the STMDF-AD with the various filters form the literature. The images tested include the Lena, Peppers, etc. Based on the results in Table 3 and Fig. 5 show that the STMDF-AD consistently outperforms several of the known median filter variants. Additionally, the filters designed for high density noise such as NAFSMF or MDBUTMF were compared. The aspects of the filters that would normally blur edges have been mitigated. The general trend is that the STMDF dominates at high noise densities and low noise density in most cases. However, since impulse noise is highly non-linear and non-additive, the results are still quite surprising.

| Filter | Lena @ 10% | Lena @ 20% | Lena @ 30% | Lena @ 40% | Lena @ 50% | Lena @ 60% | Lena @ 70% | Lena @80% | Lena @90% |
|---|---|---|---|---|---|---|---|---|---|
| Med | 33.1027 | 28.5209 | 23.5908 | 18.804 | 15.1522 | 12.3633 | 9.9962 | 8.1546 | 6.6478 |
| AMF | 37.9556 | 34.8321 | 32.7849 | 30.4298 | 28.3124 | 26.4811 | 24.4653 | 20.3704 | 13.7031 |
| MDBUTMF | 37.7178 | 34.604 | 32.6131 | 31.3683 | 30.0338 | 29.0234 | 28.1496 | 27.137 | 25.3418 |
| FIRE2n | 31.86 | 24.6061 | 19.7341 | 15.9973 | 13.1915 | 11.0473 | 9.205 | 7.7223 | 6.4728 |
| FIRE2r | 34.2373 | 29.9134 | 26.4244 | 22.7039 | 19.0685 | 15.8659 | 12.7197 | 9.8621 | 7.3686 |
| STMDF_AD | **40.6408** | **37.4805** | **35.4109** | **34.1019** | **33.9** | **32.1** | **30.7609** | **29.41** | **27.4923** |
| FESTM | 34.5911 | 32.7889 | 31.1052 | 29.3935 | 27.0312 | 23.3868 | 18.7982 | 14.0849 | 9.6617 |
| SMF | 31.0467 | 23.2544 | 17.4936 | 13.9458 | 11.6021 | 9.818 | 8.2855 | 7.1331 | 6.2116 |
| PSMF | 37.0489 | 32.1836 | 29.0489 | 24.9683 | 20.6717 | 12.2828 | 9.9384 | 8.1176 | 6.6318 |
| NAFSM | 38.816 | 35.6916 | 33.6541 | 32.3401 | 30.8753 | 29.7736 | 28.6372 | 27.1376 | 23.4373 |
| DWFM | 33.6526 | 29.3519 | 23.832 | 18.9049 | 15.1278 | 12.3176 | 9.9253 | 8.0942 | 6.6083 |

(a)

| Filter | Peppers @10% | Peppers @20% | Peppers @30% | Peppers @40% | Peppers @50% | Peppers @60% | Peppers @70% | Peppers @80% | Peppers @90% |
|---|---|---|---|---|---|---|---|---|---|
| Med | 31.6302 | 28.1805 | 23.098 | 18.7499 | 15.0541 | 12.1611 | 9.8015 | 7.9568 | 6.4647 |
| AMF | 34.9864 | 33.6469 | 31.6547 | 29.8014 | 28.0815 | 26.3943 | 24.2688 | 20.2563 | 3.6766 |
| MDBUTMF | 38.5032 | 35.3375 | 33.5738 | 32.106 | 30.8644 | 29.7692 | 28.6105 | 27.2986 | 25.5261 |
| FIRE2n | 31.5853 | 24.389 | 19.4439 | 15.9145 | 13.0497 | 10.8476 | 9.0129 | 7.5625 | 6.2995 |
| FIRE2r | 34.0725 | 29.8773 | 26.3134 | 22.6046 | 19.0393 | 15.6859 | 12.5549 | 9.7325 | 7.2237 |
| STMDF-AD | 39.1913 | **36.4405** | **34.6947** | **33.2775** | **32.1014** | **31.1756** | **30.1650** | **29.3339** | **27.5463** |
| FESTM | 33.3291 | 31.8693 | 30.5704 | 29.0457 | 26.833 | 23.6212 | 19.0386 | 14.2936 | 9.9883 |
| SMF | 31.4874 | 21.6549 | 17.3826 | 14.5223 | 12.0192 | 10.1006 | 8.4668 | 7.1712 | 6.11 |
| PSMF | 36.2691 | 31.4148 | 28.0333 | 24.438 | 15.0137 | 12.1237 | 9.7637 | 7.9323 | 6.4534 |
| NAFSM | **39.519** | 36.3048 | 34.4985 | 32.9166 | 31.6165 | 30.4287 | 28.9183 | 27.2275 | 23.669 |
| DWFM | 32.1974 | 28.949 | 23.4036 | 18.8778 | 15.0752 | 12.1434 | 9.7637 | 7.9189 | 6.4568 |

(b)

| Filter | Gold Hill @10% | Gold Hill @20% | Gold Hill @30% | Gold Hill @40% | Gold Hill @50% | Gold Hill @60% | Gold Hill @70% | Gold Hill @80% | Gold Hill @90% |
|---|---|---|---|---|---|---|---|---|---|
| Med | 30.3738 | 27.5088 | 22.9758 | 18.6167 | 15.0163 | 12.1487 | 9.9252 | 8.032 | 6.5549 |
| AMF | 34.7316 | 33.1146 | 30.7087 | 28.7857 | 27.1514 | 25.7042 | 23.828 | 19.969 | 13.7588 |
| MDBUTMF | 36.2434 | 33.0963 | 31.2686 | 29.9325 | 28.9211 | 27.971 | 27.1175 | 26.1592 | 24.8885 |
| FIRE2n | 31.0727 | 24.472 | 19.5385 | 15.9344 | 13.0972 | 10.8758 | 9.1292 | 7.6289 | 6.3973 |
| FIRE2r | 33.4551 | 29.6473 | 25.9814 | 22.2676 | 18.9262 | 15.6088 | 12.6672 | 9.7692 | 7.3897 |
| STMDF-AD | 37.1380 | **34.5137** | **32.4310** | **31.3515** | **30.3708** | **29.5245** | **28.5399** | **27.8406** | **26.4769** |
| FESTM | 31.8025 | 30.5854 | 29.4478 | 28.0615 | 26.2639 | 23.3441 | 19.2596 | 14.7352 | 10.3501 |
| SMF | 31.0092 | 22.5898 | 17.3721 | 13.9506 | 11.5167 | 9.6903 | 8.313 | 7.1101 | 6.1546 |
| PSMF | 33.9631 | 30.8805 | 27.4432 | 24.4398 | 20.5103 | 16.2378 | 9.8873 | 8.0059 | 6.5421 |
| NAFSM | **37.1957** | 33.9979 | 32.1461 | 30.7413 | 29.6955 | 28.6354 | 27.6159 | 26.3618 | 23.1123 |
| DWFM | 30.8608 | 28.133 | 23.3321 | 18.701 | 15.0454 | 12.1284 | 9.885 | 8.0038 | 6.5577 |

(c)

| Filter | Boat @10% | Boat @20% | Boat @30% | Boat @40% | Boat @50% | Boat @60% | Boat @70% | Boat @80% | Boat @90% |
|---|---|---|---|---|---|---|---|---|---|
| Med | 30.4152 | 26.9764 | 22.551 | 18.5473 | 15.0613 | 12.2387 | 9.8783 | 8.0182 | 6.5112 |
| AMF | 35.1825 | 32.5336 | 30.2059 | 28.3305 | 26.6616 | 24.8546 | 23.0974 | 19.4761 | 13.5018 |
| MDBUTMF | 34.3858 | 31.4113 | 29.3917 | 28.1948 | 27.1762 | 26.1744 | 25.2691 | 24.3889 | 23.0614 |
| FIRE2n | 30.7574 | 24.0902 | 19.3184 | 15.9015 | 13.1291 | 10.9506 | 9.0938 | 7.5984 | 6.3524 |
| FIRE2r | 32.9701 | 28.9356 | 25.3348 | 22.0402 | 18.73 | 15.6301 | 12.5034 | 9.6918 | 7.2486 |
| STMDF-AD | **35.9495** | **33.7142** | **32.2112** | **31.0295** | **30.0703** | **28.8984** | **27.7683** | **26.5792** | **24.9939** |
| FESTM | 32.3517 | 30.9514 | 29.5103 | 28.118 | 25.9633 | 22.3425 | 17.9261 | 13.2242 | 8.8629 |
| SMF | 31.4778 | 20.9695 | 17.0984 | 14.4257 | 12.1527 | 10.2746 | 8.5931 | 7.2238 | 6.1552 |
| PSMF | 33.1835 | 29.7946 | 26.7411 | 23.7352 | 20.3037 | 16.2349 | 9.8519 | 8.0013 | 6.5022 |
| NAFSM | 35.7769 | 32.8192 | 30.6852 | 29.4707 | 28.3268 | 27.2085 | 26.0275 | 24.7979 | 22.1008 |
| DWFM | 30.925 | 27.6035 | 22.8411 | 18.6702 | 15.0553 | 12.2028 | 9.8102 | 7.9426 | 6.4476 |

(d)

| Filter | Bridge @ 10% | Bridge @ 20% | Bridge @ 30% | Bridge @ 40% | Bridge @ 50% | Bridge @ 60% | Bridge @ 70% | Bridge @ 80% | Bridge @90% |
|---|---|---|---|---|---|---|---|---|---|
| Med | 25.9369 | 24.2888 | 21.2312 | 17.6424 | 14.6564 | 11.9146 | 9.7484 | 7.8996 | 6.4669 |
| AMF | 28.6338 | 27.0595 | 25.4723 | 24.0825 | 22.8558 | 21.6548 | 20.2167 | 17.7652 | 12.8858 |
| MDBUTMF | 29.9867 | 27.7344 | 26.1779 | 24.9423 | 24.056 | 23.2189 | 22.39 | 21.5142 | 20.3954 |
| FIRE2n | 28.4151 | 23.1803 | 18.808 | 15.385 | 12.8311 | 10.6818 | 8.9627 | 7.4797 | 6.2871 |
| FIRE2r | 29.6863 | 26.5333 | 23.6657 | 20.5991 | 17.8651 | 14.8851 | 12.1551 | 9.4751 | 7.1808 |
| STMDF-AD | 30.6392 | 28.4242 | **26.8421** | **26.2742** | **24.8058** | **24.1869** | **23.3125** | **22.6996** | **21.7225** |
| FESTM | 26.9034 | 26.1525 | 25.2555 | 24.3432 | 23.1328 | 21.203 | 17.9265 | 14.0184 | 9.9545 |
| SMF | 28.1935 | 21.2808 | 16.8921 | 13.9209 | 11.6066 | 9.7546 | 8.3674 | 7.067 | 6.0993 |
| PSMF | 28.8009 | 26.9044 | 24.7095 | 22.0798 | 19.2219 | 11.8825 | 9.7073 | 7.8749 | 6.4523 |
| NAFSM | **31.3195** | **28.5972** | 26.8164 | 25.5224 | 24.4558 | 23.5474 | 22.5367 | 21.46 | 19.4492 |
| DWFM | 26.1689 | 24.5403 | 21.3825 | 17.632 | 14.6405 | 11.8692 | 9.6938 | 7.8534 | 6.4496 |

(e)

| Filter | Jetplane @ 10% | Jetplane @ 20% | Jetplane @ 30% | Jetplane @ 40% | Jetplane @ 50% | Jetplane @ 60% | Jetplane @ 70% | Jetplane @ 80% | Jetplane @ 90% |
|---|---|---|---|---|---|---|---|---|---|
| Med | 34.1955 | 28.8073 | 23.3249 | 18.816 | 15.1839 | 12.1237 | 9.8855 | 8.0256 | 6.5277 |
| AMF | 39.2939 | 36.4385 | 33.9087 | 31.4531 | 29.4069 | 27.4818 | 25.6558 | 20.7555 | 13.8314 |
| MDBUTMF | 40.7622 | 38.5755 | 36.6247 | 34.7777 | 33.7884 | 32.7628 | 31.6871 | 30.5502 | 29.088 |
| FIRE2n | 31.4109 | 24.601 | 19.735 | 16.0518 | 13.2519 | 10.8654 | 9.1243 | 7.6342 | 6.3692 |
| FIRE2r | 34.3811 | 30.5359 | 26.7771 | 23.1676 | 19.4401 | 15.8516 | 12.7486 | 9.8983 | 7.3458 |
| STMDF-AD | **42.2972** | **39.9119** | **37.9374** | **36.7430** | **35.6003** | **34.5760** | **33.2587** | **32.1504** | **30.6120** |
| FESTM | 36.9321 | 35.054 | 33.0193 | 30.9321 | 27.3182 | 21.9406 | 16.7973 | 11.7913 | 7.2401 |
| SMF | 30.9657 | 20.4141 | 17.7816 | 15.4367 | 13.1405 | 10.9765 | 9.2158 | 7.6662 | 6.3757 |
| PSMF | 37.7866 | 32.3406 | 28.2301 | 18.8287 | 15.1796 | 12.1159 | 9.8653 | 8.0154 | 6.5168 |
| NAFSM | 42.2716 | 39.8171 | 37.7997 | 36.0113 | 34.7217 | 33.3641 | 32.0672 | 30.2168 | 25.3816 |
| DWFM | 35.5726 | 30.3369 | 24.0061 | 19.0915 | 15.2346 | 12.0983 | 9.7892 | 7.902 | 6.3826 |

(f)

| Filter | Tank @10% | Tank @20% | Tank @30% | Tank @40% | Tank @50% | Tank @60% | Tank @70% | Tank @80% | Tank @90% |
|---|---|---|---|---|---|---|---|---|---|
| **Med** | 30.0025 | 27.1871 | 23.1893 | 19.024 | 15.4205 | 12.5014 | 10.2183 | 8.3542 | 6.8505 |
| **AMF** | 34.1082 | 32.6832 | 30.8218 | 29.146 | 27.8142 | 25.9753 | 24.351 | 20.4458 | 13.9375 |
| **MDBUTMF** | 36.2048 | 33.0564 | 31.2977 | 30.0399 | 28.9961 | 28.1565 | 27.3451 | 26.5535 | 25.4226 |
| **FIRE2n** | 31.4894 | 24.441 | 19.8387 | 16.3195 | 13.5142 | 11.2442 | 9.4572 | 7.9443 | 6.7124 |
| **FIRE2r** | 33.6 | 29.5137 | 26.1526 | 22.8277 | 19.2651 | 15.8706 | 12.9073 | 10.0194 | 7.5928 |
| **STMDF-AD** | **38.2077** | **35.3229** | **33.4854** | **31.6116** | **30.6974** | **29.5158** | **28.4204** | **27.2103** | **26.5622** |
| **FESTM** | 31.6216 | 30.5682 | 29.5289 | 28.3502 | 26.4515 | 23.2304 | 18.7967 | 14.1962 | 9.7877 |
| **SMF** | 34.6265 | 22.014 | 16.7271 | 13.8737 | 11.5653 | 9.6675 | 8.2693 | 7.1148 | 6.2984 |
| **PSMF** | 34.929 | 31.3565 | 28.5318 | 25.0161 | 21.1666 | 12.5204 | 10.2166 | 8.3477 | 6.8458 |
| **NAFSM** | 36.8193 | 33.6768 | 31.9032 | 30.6355 | 29.5314 | 28.6312 | 27.6246 | 26.4707 | 23.6473 |
| **DWFM** | 30.3836 | 27.741 | 23.4497 | 19.1523 | 15.4202 | 12.4701 | 10.1648 | 8.2971 | 6.8156 |

(g)

| Filter | Elaine @10% | Elaine @20% | Elaine @30% | Elaine @40% | Elaine @50% | Elaine @60% | Elaine @70% | Elaine @80% | Elaine @90% |
|---|---|---|---|---|---|---|---|---|---|
| **Med** | 31.3272 | 28.3503 | 23.4363 | 18.909 | 15.4152 | 12.2336 | 10.0335 | 8.139 | 6.6519 |
| **AMF** | 35.5045 | 34.253 | 32.2521 | 30.1324 | 28.5876 | 26.8843 | 24.9864 | 20.5869 | 13.9379 |
| **MDBUTMF** | 39.9138 | 36.7861 | 34.9578 | 33.4897 | 32.4413 | 31.2611 | 30.2057 | 28.9556 | 27.1654 |
| **FIRE2n** | 31.937 | 24.7876 | 19.6886 | 16.151 | 13.3754 | 10.9735 | 9.2395 | 7.7395 | 6.4954 |
| **FIRE2r** | 34.2477 | 30.3112 | 26.7043 | 22.9129 | 19.4948 | 15.8724 | 12.8454 | 9.8421 | 7.4121 |
| **STMDF-AD** | 40.2527 | 37.5168 | **35.7253** | **34.4048** | **33.1160** | **31.7822** | **30.9592** | **30.0927** | **28.5875** |
| **FESTM** | 33.2129 | 32.0517 | 30.7888 | 29.4232 | 27.1043 | 23.036 | 18.239 | 13.3867 | 9.0827 |
| **SMF** | 31.8305 | 22.8719 | 17.4789 | 13.9509 | 11.493 | 9.6382 | 8.2987 | 7.1368 | 6.2142 |
| **PSMF** | 36.4616 | 32.3524 | 28.7944 | 25.0043 | 21.0951 | 16.4339 | 10.009 | 8.1197 | 6.6415 |
| **NAFSM** | **40.7298** | **37.5434** | 35.6673 | 34.1886 | 32.9855 | 31.6996 | 30.4155 | 28.8028 | 24.5467 |
| **DWFM** | 32.1092 | 29.1776 | 23.8458 | 19.0844 | 15.4385 | 12.1877 | 9.9636 | 8.0631 | 6.5831 |

(h)

| Filter | Truck @10% | Truck @20% | Truck @30% | Truck @40% | Truck @50% | Truck @60% | Truck @70% | Truck @80% | Truck @90% |
|---|---|---|---|---|---|---|---|---|---|
| **Med** | 32.1546 | 28.8389 | 23.8224 | 19.2901 | 15.5132 | 12.6205 | 10.2793 | 8.4576 | 6.9474 |
| **AMF** | 36.3487 | 34.6487 | 32.7765 | 30.9503 | 29.1538 | 27.3257 | 25.2459 | 20.9441 | 14.3348 |
| **MDBUTMF** | 37.7222 | 34.7338 | 32.8943 | 31.5181 | 30.5048 | 29.6064 | 28.7801 | 27.9462 | 26.7139 |
| **FIRE2n** | 31.9867 | 24.8464 | 20.0254 | 16.3969 | 13.524 | 11.3183 | 9.4936 | 8.039 | 6.7778 |
| **FIRE2r** | 34.4967 | 30.4041 | 26.8902 | 23.2645 | 19.413 | 16.1164 | 13.0723 | 10.1948 | 7.7374 |
| **STMDF-AD** | **40.8972** | **37.6033** | **35.4602** | **33.2920** | **33.1305** | **31.9333** | **30.6792** | **29.3114** | **28.2050** |
| **FESTM** | 33.8763 | 32.4549 | 31.0904 | 29.6947 | 27.7566 | 24.5713 | 20.1599 | 15.6916 | 11.2521 |
| **SMF** | 35.0657 | 24.8048 | 18.594 | 13.1164 | 10.5324 | 9.0084 | 7.9454 | 7.0979 | 6.3562 |
| **PSMF** | 36.9988 | 33.3732 | 29.9112 | 26.1816 | 21.4643 | 12.6218 | 10.2706 | 8.4476 | 6.9416 |
| **NAFSM** | 38.6716 | 35.6235 | 33.8183 | 32.3636 | 31.3197 | 30.3379 | 29.3057 | 27.9503 | 24.6171 |
| **DWFM** | 32.5286 | 29.2279 | 23.9675 | 19.3083 | 15.4854 | 12.5822 | 10.2358 | 8.4287 | 6.9528 |

(i)

**Table 3**. PSNR values for the filtered (a) Lena (b) Peppers (c) Gold Hill (d) Boat (e) Bridge (f) Jetplane (g) Tank (h)

Elaine (i) Truck image corrupted by varying levels of FVIN noise

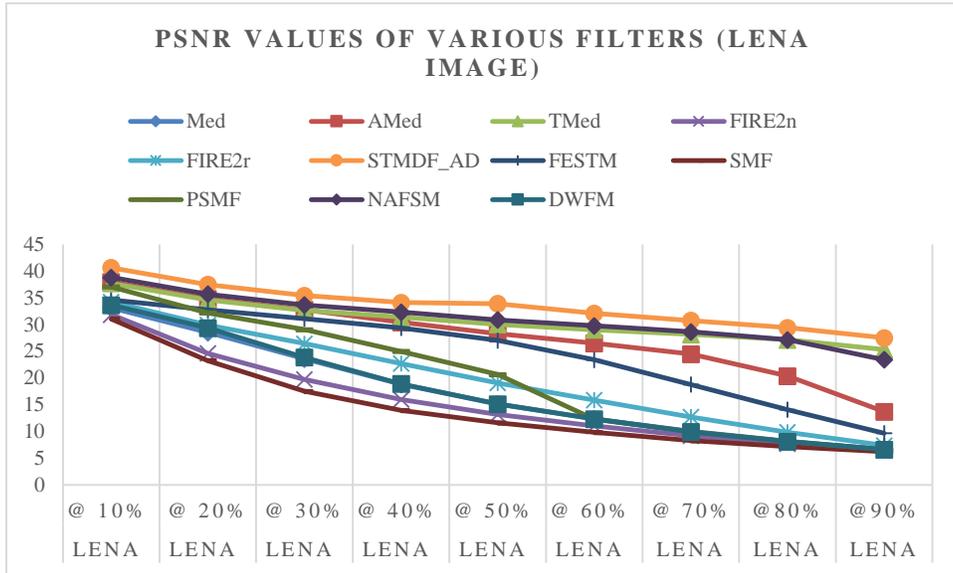

(a)

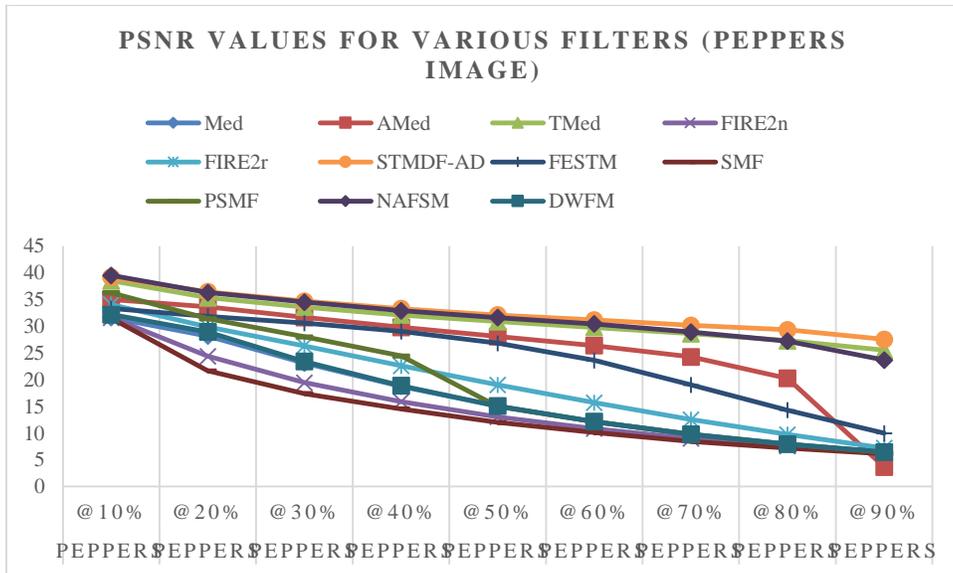

(b)

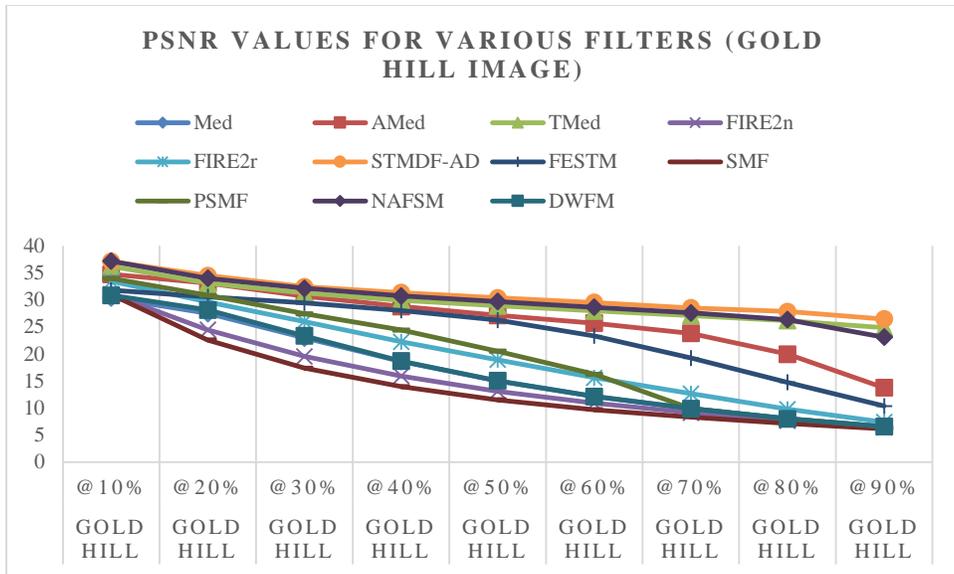

(c)

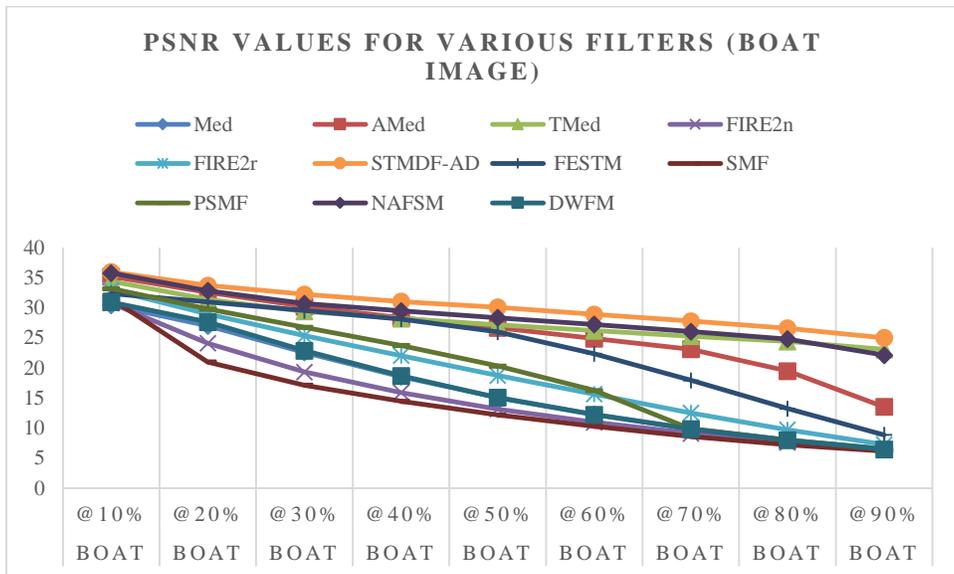

(d)

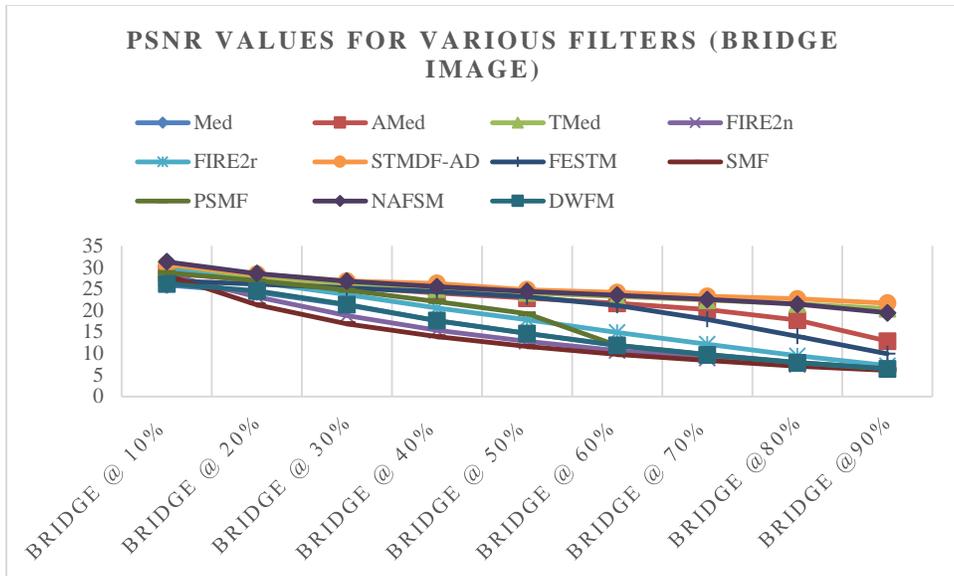

(e)

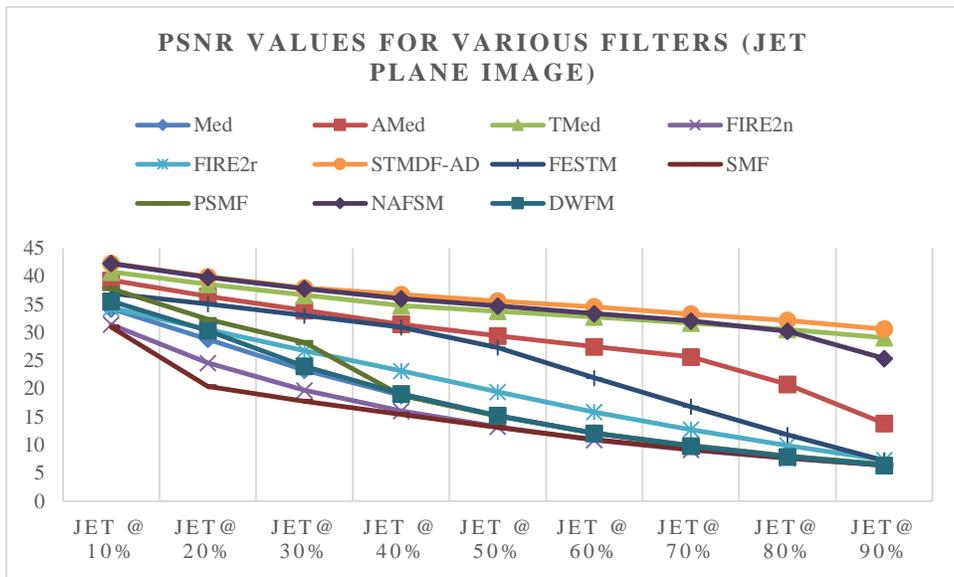

(f)

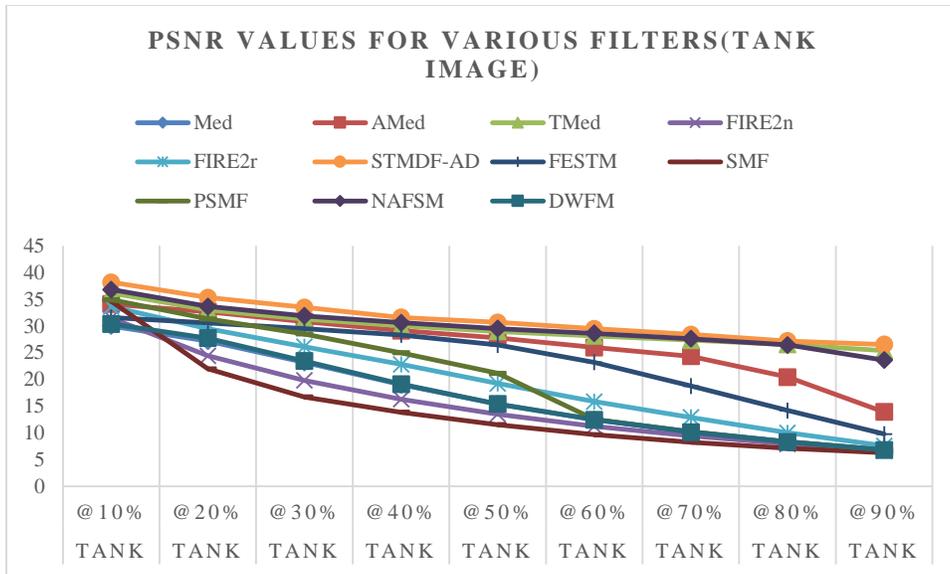

(g)

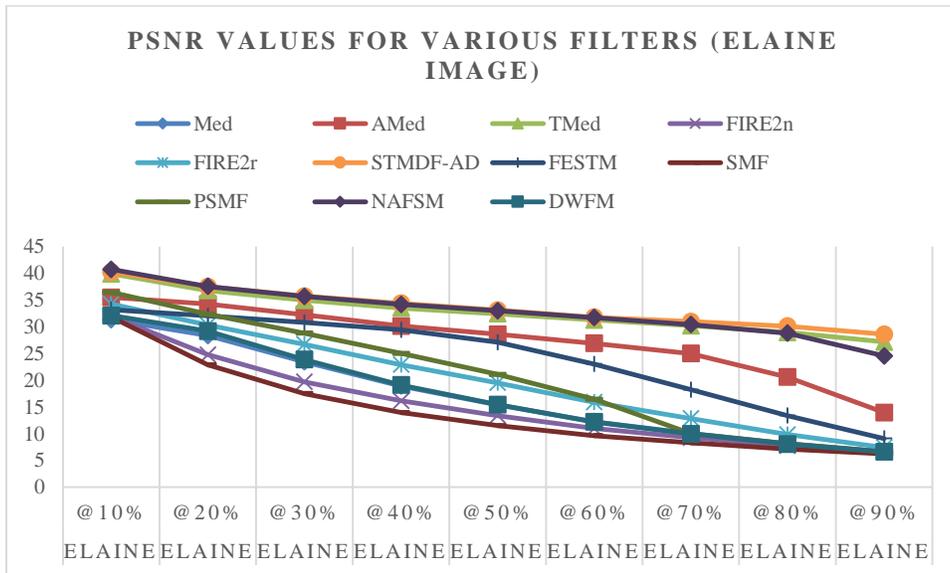

(h)

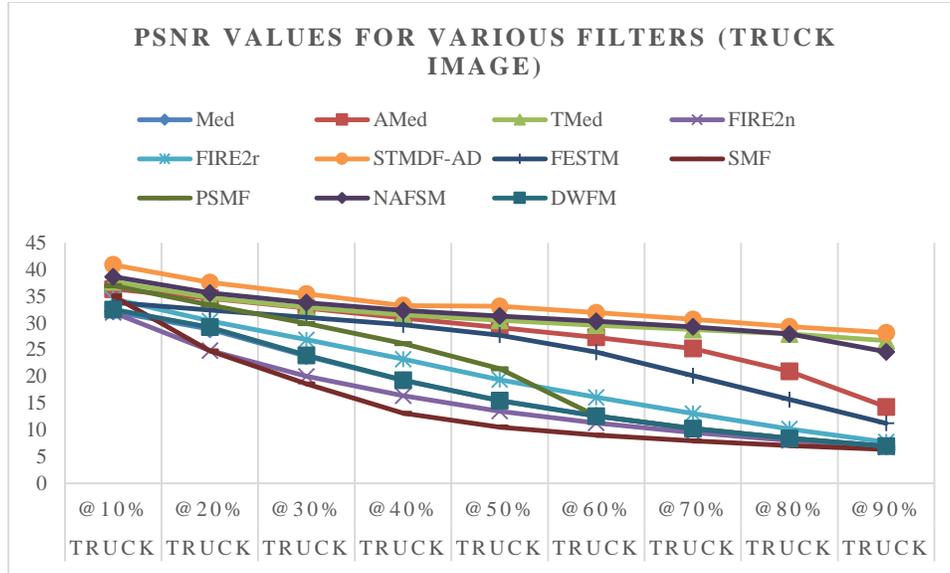

(i)

**Fig. 5**. PSNR values for the filtered (a) Lena (b) Peppers (c) Gold Hill (d) Boat (e) Bridge (f) Jetplane (g) Tank (h) Elaine (i) Truck image corrupted by varying levels of FVIN noise

The results in Table 5 once more indicates the complete domination of the proposed algorithm at even high noise densities (95%) over the other compared filters, especially the ones suited to high density noise.

| Filter | Lena @ 10% | Lena @ 20% | Lena @ 30% | Lena @ 40% | Lena @ 50% | Lena @ 60% | Lena @ 70% | Lena @80% | Lena @90% |
|---|---|---|---|---|---|---|---|---|---|
| **Med** | 2.8125 | 3.5894 | 5.3008 | 9.4641 | 17.3464 | 29.5304 | 47.6341 | 70.6258 | 97.9845 |
| **AMF** | 0.9699 | 1.3111 | 1.812 | 2.465 | 3.3512 | 4.3998 | 5.841 | 9.2384 | 24.5079 |
| **MDBUTMF** | 0.5651 | 1.1505 | 1.7702 | 2.3665 | 3.064 | 3.777 | 4.5245 | 5.4327 | 7.0298 |
| **FIRE2n** | 1.5068 | 3.9108 | 8.258 | 15.6119 | 26.3763 | 40.2507 | 58.2288 | 79.2356 | 102.8278 |
| **FIRE2r** | 1.3681 | 2.8402 | 4.7136 | 7.4604 | 12.0432 | 19.4367 | 32.4942 | 54.4271 | 87.5895 |
| **STMDF-AD** | **0.4248** | **0.8768** | **1.3686** | **1.8617** | **2.4783** | **2.6302** | **3.8897** | **4.7378** | **5.9844** |
| **FESTM** | 2.5696 | 3.1186 | 3.8006 | 4.5864 | 5.6868 | 7.6027 | 12.1309 | 23.9798 | 54.1608 |
| **SMF** | 1.1527 | 4.4487 | 12.5529 | 24.2584 | 37.9379 | 53.6621 | 72.1496 | 90.7215 | 109.1785 |
| **PSMF** | 0.5843 | 1.2157 | 2.0827 | 3.7136 | 7.3531 | 29.8466 | 48.4943 | 71.6735 | 98.7092 |
| **NAFSM** | 0.5037 | 1.0231 | 1.5868 | 2.132 | 2.792 | 3.4579 | 4.2123 | 5.2067 | 7.5689 |
| **DWFM** | 2.7743 | 3.4679 | 5.1802 | 9.3155 | 17.3878 | 29.7703 | 48.3054 | 71.5431 | 98.853 |

(a)

| Filter | Peppers @10% | Peppers @20% | Peppers @30% | Peppers @40% | Peppers @50% | Peppers @60% | Peppers @70% | Peppers @80% | Peppers @90% |
|---|---|---|---|---|---|---|---|---|---|
| **Med** | 3.3749 | 4.0596 | 5.788 | 9.6379 | 17.4401 | 29.9491 | 48.0263 | 71.1551 | 98.4177 |
| **AMF** | 1.7333 | 1.8577 | 2.2217 | 2.7951 | 3.5939 | 4.5503 | 5.9577 | 9.2714 | 24.1218 |
| **MDBUTMF** | 0.5346 | 1.0887 | 1.6384 | 2.2398 | 2.8539 | 3.5416 | 4.2944 | 5.2707 | 6.7938 |
| **FIRE2n** | 1.5046 | 3.9571 | 8.3228 | 15.3432 | 26.2129 | 40.499 | 58.4639 | 9.1549 | 102.9972 |
| **FIRE2r** | 1.358 | 2.8471 | 4.6705 | 7.3413 | 11.8066 | 19.4558 | 32.4752 | 4.1012 | 87.299 |
| **STMDF-AD** | 0.5695 | 1.1342 | 1.6975 | 2.6276 | 3.4453 | 3.8265 | 4.8752 | 5.1978 | **6.3560** |
| **FESTM** | 3.0628 | 3.6136 | 4.2184 | 4.9803 | 6.0304 | 7.7686 | 12.1054 | 23.4441 | 51.0838 |
| **SMF** | 1.0818 | 5.9605 | 12.9354 | 21.903 | 34.6948 | 49.688 | 67.8128 | 87.6864 | 107.947 |
| **PSMF** | 0.6466 | 1.4474 | 2.5005 | 4.1881 | 16.5319 | 29.6266 | 48.3915 | 71.7782 | 98.9256 |
| **NAFSM** | **0.4959** | **1.0051** | **1.5075** | **2.0725** | **2.6564** | **3.308** | **4.0835** | **5.1419** | 7.3529 |
| **DWFM** | 3.3222 | 3.9444 | 5.6405 | 9.4811 | 17.3982 | 30.0925 | 48.4829 | 71.8637 | 98.8587 |

(b)

| Filter | Gold Hill @10% | Gold Hill @20% | Gold Hill @30% | Gold Hill @40% | Gold Hill @50% | Gold Hill @60% | Gold Hill @70% | Gold Hill @80% | Gold Hill @90% |
|---|---|---|---|---|---|---|---|---|---|
| **Med** | 4.2472 | 5.0012 | 6.8065 | 10.8842 | 18.6689 | 31.1852 | 48.5912 | 71.7952 | 98.49 |
| **AMF** | 1.4532 | 1.8598 | 2.5104 | 3.3342 | 4.3494 | 5.5558 | 7.1987 | 10.8972 | 25.2825 |
| **MDBUTMF** | 0.7968 | 1.6145 | 2.4436 | 3.2965 | 4.1604 | 5.0735 | 6.0207 | 7.1712 | 8.7866 |
| **FIRE2n** | 1.6601 | 4.1262 | 8.5914 | 15.8461 | 26.6661 | 41.0512 | 58.5167 | 79.4677 | 102.7248 |
| **FIRE2r** | 1.5028 | 3.0435 | 5.0134 | 7.9659 | 12.4717 | 20.2472 | 33.0117 | 54.9444 | 86.3491 |
| **STMDF-AD** | **0.7052** | **1.3712** | 2.2790 | 3.0387 | 3.8744 | **4.6311** | 5.6119 | **6.6332** | **7.8370** |
| **FESTM** | 3.8372 | 4.4313 | 5.103 | 5.9407 | 6.9998 | 8.7623 | 12.8027 | 23.3685 | 50.187 |
| **SMF** | 1.2888 | 5.1993 | 12.9434 | 24.2575 | 38.3142 | 54.235 | 71.0251 | 89.8036 | 108.67 |
| **PSMF** | 1.0222 | 1.8142 | 2.9847 | 4.6623 | 8.0148 | 15.5688 | 48.8637 | 72.4581 | 99.0538 |
| **NAFSM** | 0.7169 | 1.4545 | **2.2081** | **2.994** | **3.7956** | 4.6741 | **5.6072** | 6.8224 | 9.272 |
| **DWFM** | 4.1995 | 4.9017 | 6.648 | 10.7712 | 18.6121 | 31.3672 | 49.0533 | 72.3927 | 98.7938 |

(c)

| Filter | Boat @10% | Boat @20% | Boat @30% | Boat @40% | Boat @50% | Boat @60% | Boat @70% | Boat @80% | Boat @90% |
|---|---|---|---|---|---|---|---|---|---|
| **Med** | 3.7314 | 4.6318 | 6.5804 | 10.5839 | 18.129 | 30.4602 | 48.5204 | 71.3438 | 98.4811 |
| **AMF** | 1.2673 | 1.7122 | 2.3814 | 3.1892 | 4.216 | 5.5024 | 7.196 | 10.9693 | 25.7975 |
| **MDBUTMF** | 0.8146 | 1.6338 | 2.5132 | 3.3475 | 4.226 | 5.1757 | 6.1681 | 7.3423 | 9.1141 |
| **FIRE2n** | 1.7474 | 4.3368 | 8.8813 | 15.9682 | 26.5491 | 40.6269 | 58.693 | 79.5176 | 102.9133 |
| **FIRE2r** | 1.5842 | 3.2023 | 5.303 | 8.214 | 12.8423 | 20.4607 | 33.8505 | 55.6071 | 87.8882 |
| **STMDF - AD** | 0.7491 | **1.3830** | **2.0328** | **2.7155** | **3.4055** | **4.3090** | **5.2925** | **6.4474** | **7.9950** |
| **FESTM** | 3.2586 | 3.8488 | 4.555 | 5.372 | 6.5084 | 8.6402 | 13.5597 | 26.6855 | 59.3964 |
| **SMF** | 1.1197 | 6.7716 | 13.7878 | 22.5186 | 34.0741 | 48.4934 | 66.656 | 87.0572 | 107.7068 |
| **PSMF** | 1.0979 | 1.9554 | 3.1801 | 4.9583 | 8.1556 | 15.4257 | 48.6142 | 71.702 | 98.8604 |
| **NAFSM** | **0.7114** | 1.4271 | 2.2079 | 2.9487 | 3.7661 | 4.6441 | 5.6546 | 6.9152 | 9.4466 |
| **DWFM** | 3.6764 | 4.5035 | 6.4194 | 10.4254 | 18.1326 | 30.6521 | 49.1487 | 72.3784 | 99.6529 |

(d)

| Filter | Bridge @10% | Bridge @20% | Bridge @30% | Bridge @40% | Bridge @50% | Bridge @60% | Bridge @70% | Bridge @80% | Bridge @90% |
|---|---|---|---|---|---|---|---|---|---|
| Med | 7.3923 | 8.4446 | 10.6144 | 15.1312 | 22.4358 | 34.7467 | 51.7067 | 73.9766 | 99.0583 |
| AMF | 2.9055 | 3.5607 | 4.6483 | 6.0311 | 7.6397 | 9.6918 | 12.353 | 16.9798 | 32.1718 |
| MDBUTMF | 1.7154 | 3.1709 | 4.6627 | 6.1698 | 7.6541 | 9.2071 | 10.8724 | 12.8413 | 15.4789 |
| FIRE2n | 2.8113 | 5.7018 | 10.6868 | 18.4476 | 28.8274 | 43.1165 | 60.3597 | 81.0394 | 103.137 |
| FIRE2r | 2.6216 | 4.5721 | 7.0929 | 10.6458 | 15.5358 | 23.8751 | 36.9649 | 58.4438 | 88.6215 |
| STMDF-AD | 1.4334 | **2.7127** | **3.9777** | **4.5839** | 7.5363 | 9.0036 | 10.6044 | **12.0269** | **14.6781** |
| FESTM | 6.5273 | 7.266 | 8.1282 | 9.1531 | 10.3587 | 12.362 | 16.7488 | 27.4376 | 53.5846 |
| SMF | 2.0075 | 6.6713 | 14.3818 | 24.5498 | 37.5339 | 52.9196 | 69.3644 | 88.8516 | 107.6434 |
| PSMF | 2.3306 | 3.576 | 5.2279 | 7.6675 | 11.4076 | 33.4184 | 51.4047 | 74.153 | 99.4896 |
| NAFSM | **1.4298** | 2.7888 | 4.2035 | 5.5951 | **7.0577** | **8.5622** | **10.2764** | 12.3344 | 15.932 |
| DWFM | 7.3074 | 8.3293 | 10.4821 | 15.1006 | 22.456 | 35.0012 | 52.2575 | 74.7082 | 99.566 |

(e)

| Filter | Jetplane @ 10% | Jetplane @ 20% | Jetplane @ 30% | Jetplane @ 40% | Jetplane @ 50% | Jetplane @ 60% | Jetplane @ 70% | Jetplane @ 80% | Jetplane @ 90% |
|---|---|---|---|---|---|---|---|---|---|
| Med | 1.6999 | 2.1968 | 3.7501 | 7.5088 | 14.9427 | 28.2407 | 46.1069 | 69.4648 | 97.3867 |
| AMF | 0.8916 | 0.9355 | 1.1214 | 1.4409 | 1.8528 | 2.4392 | 3.2013 | 6.0166 | 20.4056 |
| MDBUTMF | 0.2828 | 0.5408 | 0.8241 | 1.1364 | 1.4278 | 1.7335 | 2.0745 | 2.4691 | **3.0664** |
| FIRE2n | 1.448 | 3.6764 | 7.7893 | 14.7517 | 25.0617 | 40.1547 | 57.6106 | 78.4602 | 102.5338 |
| FIRE2r | 1.2985 | 2.5943 | 4.265 | 6.6798 | 10.845 | 18.5225 | 31.4684 | 52.7336 | 86.3684 |
| STMDF-AD | 0.2684 | 0.5186 | 0.7892 | 1.0605 | 1.3441 | 1.6689 | 2.1212 | 2.4849 | 3.1874 |
| FESTM | 1.5799 | 1.8817 | 2.3539 | 2.9794 | 4.0056 | 6.4185 | 12.3501 | 29.074 | 72.8688 |
| SMF | 0.9537 | 7.7057 | 12.8097 | 19.376 | 28.9139 | 42.4418 | 59.1248 | 79.6127 | 103.1506 |
| PSMF | 0.3274 | 0.7174 | 1.3689 | 6.6092 | 14.2706 | 27.8234 | 46.2323 | 69.6391 | 97.8809 |
| NAFSM | **0.2522** | **0.4906** | **0.7537** | **1.0318** | **1.3179** | **1.629** | **1.9705** | **2.4386** | 3.8593 |
| DWFM | 1.6596 | 2.0564 | 3.5141 | 7.2354 | 14.851 | 28.4103 | 46.9526 | 71.001 | 99.6992 |

(f)

| Filter | Tank @10% | Tank @20% | Tank @30% | Tank @40% | Tank @50% | Tank @60% | Tank @70% | Tank @80% | Tank @90% |
|---|---|---|---|---|---|---|---|---|---|
| Med | 4.9559 | 5.7361 | 7.4633 | 11.3951 | 19.049 | 31.6686 | 49.3562 | 72.1634 | 99.1763 |
| AMF | 1.7643 | 2.1838 | 2.8415 | 3.6791 | 4.7078 | 6.0771 | 7.7376 | 11.4357 | 26.5359 |
| MDBUTMF | 0.8841 | 1.8052 | 2.7083 | 3.6148 | 4.5517 | 5.4949 | 6.5231 | 7.6413 | 9.2151 |
| FIRE2n | 1.6208 | 4.2371 | 8.6382 | 15.7335 | 26.4167 | 40.9927 | 58.7311 | 79.8464 | 103.0507 |
| FIRE2r | 1.4801 | 3.0907 | 5.0658 | 7.8085 | 12.4395 | 20.4068 | 33.3828 | 55.6825 | 88.0801 |
| STMDF-AD | **0.6772** | **1.3648** | **2.0720** | **2.9372** | **3.8250** | **4.8844** | **6.0007** | **7.3552** | **8.4341** |
| FESTM | 4.329 | 4.8585 | 5.4635 | 6.2168 | 7.2345 | 9.0917 | 13.6646 | 25.5152 | 55.821 |
| SMF | 0.8392 | 5.5012 | 14.5808 | 25.1813 | 39.2437 | 56.6947 | 74.9126 | 94.4931 | 111.949 |
| PSMF | 0.9472 | 1.7628 | 2.7821 | 4.3859 | 7.3788 | 29.9134 | 48.3657 | 71.832 | 99.2161 |
| NAFSM | 0.8165 | 1.6703 | 2.5114 | 3.3545 | 4.2493 | 5.1518 | 6.1997 | 7.4147 | 9.7489 |
| DWFM | 4.8967 | 5.6128 | 7.3225 | 11.2253 | 19.0446 | 31.8481 | 49.895 | 73.043 | 99.9878 |

(g)

| Filter | Elaine @10% | Elaine @20% | Elaine @30% | Elaine @40% | Elaine @50% | Elaine @60% | Elaine @70% | Elaine @80% | Elaine @90% |
|---|---|---|---|---|---|---|---|---|---|
| **Med** | 4.2704 | 4.8658 | 6.5155 | 10.4514 | 17.6577 | 31.0104 | 48.3497 | 71.6879 | 98.4158 |
| **AMF** | 1.5714 | 1.86 | 2.3711 | 3.074 | 3.9187 | 5.005 | 6.4295 | 9.8205 | 24.213 |
| **MDBUTMF** | 0.5769 | 1.1642 | 1.764 | 2.3879 | 3.0112 | 3.7129 | 4.4602 | 5.3896 | 6.7497 |
| **FIRE2n** | 1.4688 | 3.838 | 8.2626 | 15.3089 | 25.7198 | 40.8155 | 58.2919 | 79.4022 | 102.6757 |
| **FIRE2r** | 1.3381 | 2.7735 | 4.6008 | 7.3093 | 11.5488 | 19.3666 | 32.1332 | 54.8584 | 86.8807 |
| **STMDF-AD** | 0.5638 | 1.1021 | **1.6482** | 2.2397 | 2.8903 | 3.6962 | 5.1324 | 5.5921 | **6.3919** |
| **FESTM** | 3.7812 | 4.209 | 4.7668 | 5.4173 | 6.3727 | 8.3545 | 13.1911 | 26.2434 | 58.2226 |
| **SMF** | 1.12 | 4.9036 | 12.6834 | 24.3911 | 39.0319 | 55.4838 | 72.417 | 91.0257 | 109.4336 |
| **PSMF** | 0.6698 | 1.4517 | 2.5065 | 4.1221 | 7.1139 | 14.576 | 48.1992 | 72.0024 | 98.8315 |
| **NAFSM** | **0.5386** | **1.0852** | 1.651 | **2.2366** | **2.8419** | **3.5257** | **4.2842** | **5.2547** | 7.3813 |
| **DWFM** | 4.2099 | 4.7483 | 6.3401 | 10.2372 | 17.5995 | 31.2792 | 48.9937 | 72.7459 | 99.6708 |

(h)

| Filter | Truck @10% | Truck @20% | Truck @30% | Truck @40% | Truck @50% | Truck @60% | Truck @70% | Truck @80% | Truck @90% |
|---|---|---|---|---|---|---|---|---|---|
| **Med** | 3.5255 | 4.1832 | 5.8861 | 9.7584 | 17.5708 | 30.0494 | 48.1125 | 70.6795 | 97.7899 |
| **AMF** | 1.2452 | 1.5778 | 2.0936 | 2.7588 | 3.5999 | 4.6967 | 6.141 | 9.5321 | 23.6759 |
| **MDBUTMF** | 0.6894 | 1.3856 | 2.0907 | 2.8195 | 3.5426 | 4.3076 | 5.1022 | 5.99 | 7.3113 |
| **FIRE2n** | 1.4907 | 3.9059 | 8.2236 | 15.3216 | 26.1699 | 40.3531 | 58.3762 | 78.8834 | 102.3647 |
| **FIRE2r** | 1.3504 | 2.7966 | 4.6086 | 7.2137 | 11.7628 | 19.2232 | 32.0794 | 53.9943 | 86.0291 |
| **STMDF-AD** | **0.4770** | **0.9890** | **1.5557** | **2.1634** | **2.6593** | **3.3708** | **4.6142** | **5.3278** | **6.5461** |
| **FESTM** | 3.1312 | 3.6248 | 4.2284 | 4.9609 | 5.9419 | 7.6539 | 11.6835 | 21.6414 | 47.5109 |
| **SMF** | 0.6969 | 3.2236 | 9.7014 | 27.8904 | 47.2536 | 64.4859 | 80.2095 | 95.7251 | 111.7216 |
| **PSMF** | 0.6565 | 1.2717 | 2.1413 | 3.4537 | 6.4147 | 28.8634 | 47.5109 | 70.5578 | 97.8903 |
| **NAFSM** | 0.6198 | 1.2499 | 1.8797 | 2.5543 | 3.2225 | 3.9405 | 4.7236 | 5.7003 | 7.7499 |
| **DWFM** | 3.4746 | 4.0975 | 5.7795 | 9.6788 | 17.6042 | 30.2573 | 48.5947 | 71.3011 | 98.096 |

(i)

**Table 4**. MAE values for the filtered (a) Lena (b) Peppers (c) Gold Hill (d) Boat (e) Bridge (f) Jetplane (g) Tank (h) Elaine (i) Truck image corrupted by varying levels of FVIN noise

| Filter | Chip @95% | Moon @95% | Lena @95% | Boat @95% | Peppers @95% | GoldHill @95% | Bridge @95% | Barbara @95% | Jetplane @95% | Truck @95% |
|---|---|---|---|---|---|---|---|---|---|---|
| **Med** | 5.2314 | 6.3761 | 6.021 | 5.891 | 5.8148 | 5.8934 | 5.7928 | 5.812 | 5.8685 | 6.2952 |
| **AMF** | 8.9291 | 10.382 | 10.1021 | 9.8221 | 9.8378 | 9.9121 | 9.4883 | 9.7172 | 9.9021 | 10.3407 |
| **MDBUTMF** | 21.1781 | 25.3215 | 23.4188 | 21.3852 | 23.181 | 22.8706 | 19.042 | 20.3106 | 25.9447 | 25.0701 |
| **FIRE2n** | 5.2628 | 6.3327 | 5.9415 | 5.829 | 5.7529 | 5.8265 | 5.7206 | 5.732 | 5.8186 | 6.2161 |
| **FIRE2r** | 5.6461 | 6.7374 | 6.3681 | 6.233 | 6.1689 | 6.1752 | 6.0619 | 6.1031 | 6.2322 | 6.5983 |
| **STMDF-AD** | **24.8560** | **26.2721** | **25.464** | **23.1929** | **25.2743** | **23.9935** | **20.4600** | **21.5561** | **28.7487** | **26.7376** |
| **FESTM** | 5.7666 | 7.7999 | 7.6282 | 6.7953 | 7.8678 | 8.3424 | 7.9873 | 7.8025 | 5.1746 | 9.2241 |
| **SMF** | 5.3125 | 6.0633 | 5.8117 | 5.7218 | 5.6509 | 5.706 | 5.6111 | 5.6368 | 5.7951 | 6.0175 |
| **PSMF** | 5.2115 | 6.3749 | 6.0137 | 5.8867 | 5.81 | 5.8872 | 5.7864 | 5.8033 | 5.8636 | 6.2927 |
| **NAFSM** | 15.4329 | 17.8306 | 17.0449 | 16.4001 | 16.6602 | 16.7081 | 15.295 | 15.9605 | 17.2913 | 17.4006 |
| **DWFM** | 5.2516 | 6.36 | 5.9988 | 5.8485 | 5.8213 | 5.9133 | 5.8072 | 5.8131 | 5.7251 | 6.3251 |

(a)

| Filter | Chip @95% | Moon @95% | Lena @95% | Boat @95% | Peppers @95% | GoldHill @95% | Bridge @95% | Barbara @95% | Jetplane @95% | Truck @95% |
|---|---|---|---|---|---|---|---|---|---|---|
| **Med** | 114.2798 | 112.7003 | 112.3466 | 112.766 | 113.2729 | 113.2921 | 113.6451 | 113.4659 | 112.7317 | 112.5882 |
| **AMF** | 51.241 | 49.6793 | 48.2025 | 50.6212 | 49.339 | 50.2664 | 56.4144 | 52.9575 | 46.3373 | 49.0485 |
| **MDBUTMF** | 7.7863 | 8.8244 | 8.9429 | 11.486 | 9.0745 | 11.0859 | 18.697 | 14.6473 | 4.5513 | 8.9322 |
| **FIRE2n** | 114.9241 | 114.506 | 114.8477 | 114.9192 | 115.4342 | 115.4344 | 115.6781 | 115.7186 | 115.0038 | 115.0504 |
| **FIRE2r** | 107.4686 | 106.2299 | 106.3822 | 107.0383 | 106.9891 | 108.2397 | 108.928 | 108.3514 | 106.8399 | 107.1618 |
| **STMDF-AD** | **5.6351** | **8.1911** | **7.4677** | **9.8832** | **7.9239** | **10.1213** | **16.6855** | **13.0784** | **3.7888** | **7.8772** |
| **FESTM** | 97.4034 | 84.0067 | 81.7432 | 90.3358 | 77.9901 | 74.6526 | 78.0441 | 79.4866 | 113.4079 | 71.2645 |
| **SMF** | 114.0967 | 120.6731 | 118.3374 | 117.6751 | 118.2608 | 118.6601 | 118.347 | 118.197 | 115.8823 | 119.8287 |
| **PSMF** | 115.8352 | 112.7209 | 112.7417 | 113.0028 | 113.5493 | 113.6463 | 113.901 | 113.879 | 113.0009 | 112.672 |
| **NAFSM** | 15.7147 | 15.1747 | 15.3445 | 17.9326 | 16.0467 | 17.5691 | 25.037 | 20.9789 | 11.8666 | 15.5563 |
| **DWFM** | 114.132 | 113.1506 | 112.9243 | 113.6319 | 113.398 | 113.2139 | 113.6116 | 113.6368 | 115.1898 | 112.4404 |

(b)

**Table 5**. (a) PSNR and (b) MAE values for the test images corrupted with 95% FVIN and filtered with the various filters (a) Lena (b) Peppers (c) Gold Hill (d) Boat (e) Bridge (f) Jetplane (g) Baboon (h) Barbara (i) Truck (j) Moon (k) Chip image corrupted by varying levels of FVIN noise

The fuzzy filters to be compared against the proposed filter include the standard median filter (Med), the recursive (FIREr) and non-recursive (FIREn) Fuzzy FIRE filters [18], Fuzzy Estimate Select Type Median Filter (FESTMF) [38] and Directional Weighted Fuzzy Median Filter (DWFMF) [39]. Based on experiments, most of the Fuzzy filters break down around the 30% to 50% noise density region. The STMDF-AD and NAFSMF algorithms yield the best qualitative results, which correspond to the two consistently highest PSNR values in Table 5.

The results from Table 5 show that the performance of the NAFSMF drops off relatively faster than that of the STMDF-AD with increasing noise density, especially around the 80% to 90% mark. This is due to the cautious nature of the STMDF-AD, which is incremental in its filtering (in addition to the entropy guided threshold) to withstand the degrading effects of the high noise levels. Comparing the visual results from Fig. 6 at impulse noise density of 90%, only the AMF, MDBUTMF (TMF), STMDF-AD, NAFSMF successfully filtered the noise. This is to be expected since the majority of these algorithms incorporate the switching scheme-based impulse noise detection methods in addition to iterative operation. The Switching MF and the PSMF also utilize switching but are only effective at relatively low noise levels. The Fuzzy and AMF filters are most effective at low to medium to moderately increased noise levels.

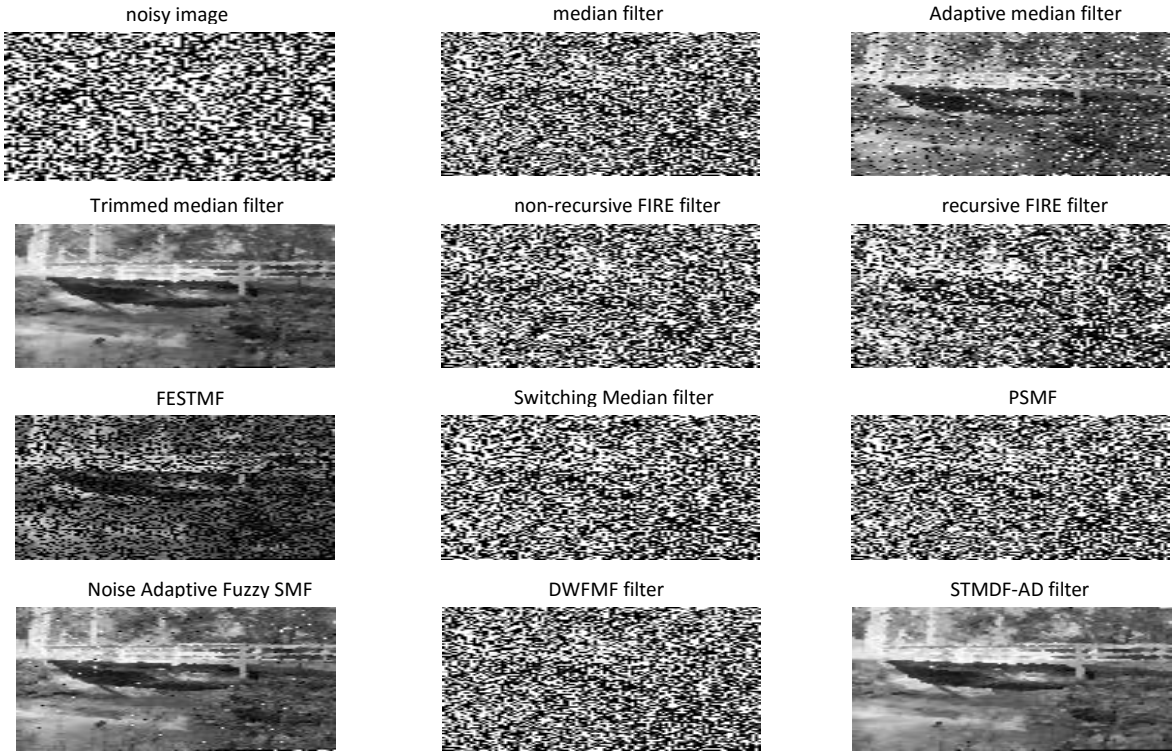

**Fig. 6** Visual performance for filters denoising Bridge image corrupted with salt and pepper impulse noise (density=90%) for the various non-Fuzzy and Fuzzy Median Filters compared with the STMDF

In summary, the MDBUTMF, NAFSMF, STMDF-AD, AMF, PSMF and Fuzzy FIRE filters dominate at medium noise density levels. At high noise levels, mostly only the MDBUTMF, STMDF-AD and NAFSMF dominate. The AMF appears to be the least robust of the adaptive and iterative filters and the STMDF-AD is the highest performing filter at very high noise density levels.

## 3.2 Comparisons with other contemporary high density noise filters

In Fig. 7, we show a sample of images corrupted with 90% salt and pepper impulse noise processed with the STMDF–AD filter. Visually, it can be seen that the filter recovers most of the original signal amidst such high level of noise even when the image signal is undiscernible to the human eye.

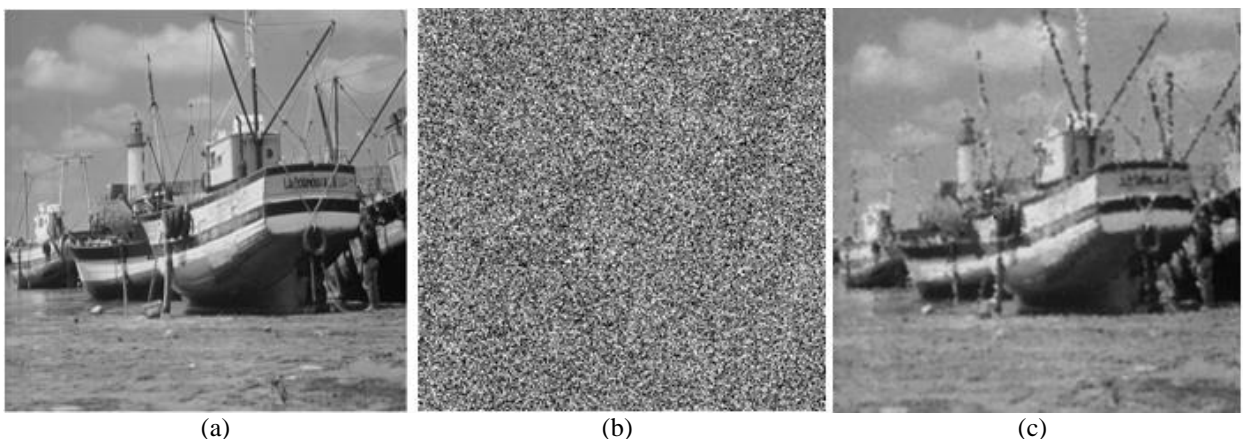

|        (a)        |        (b)        |        (c)        |

**Fig. 7** (a) Original Boat image (b) image corrupted with salt and pepper noise (ND = 90%) (c) filtered image using STMDF-AD

The other relatively recent high density noise filters from the literature compared with proposed algorithm include the Decision-Based Algorithm (DBA), the Median Filters with Regularization Chan, et al, the Fuzzy Cellular Automata (FCA) algorithm, Iterated Truncated Mean Filter (ITMF), DDBSM, ISM, SAWM, Edge Preserving Algorithm (EPA), Open and Close Filter (OCS), Linear Predictive Coding Switching Median Filter (LPC-SMF), DAM, CM and FEMF in addition to NASSM and NAFSMF and MDBUTMF. The results are presented in Table 6 to 8 and Figs 8 to 11. Based on the visual and quantitative results, the STMDF-AD once more dominates at high to extreme noise density levels of salt and pepper impulse noise.

**Table 6** Comparison of Lena and Bridge images filtered with various filters from [38] and STMDF-AD

| Algorithm | PSNR (Lena) ND (50%) | PSNR (Lena) ND (60%) | PSNR (Lena) ND (70%) | SSIM (Bridge) ND (70%) | SSIM (Lena) ND (70%) | PSNR (Bridge) ND (50%) | PSNR (Bridge) ND (60%) | PSNR (Bridge) ND (70%) |
|---|---|---|---|---|---|---|---|---|
| PSM | 24.7 | 24.4 | 19.4 | 0.7 | 0.55 | 22.2 | 19.8 | 17 |
| DDBSM | 24.5 | 21.8 | 17.4 | 0.32 | 0.27 | 22.4 | 19.2 | 15.9 |
| ISM | 27.7 | 24.9 | 23.3 | 0.68 | 0.45 | 23.1 | 22.7 | 20.1 |
| NASM | 28.9 | 24.6 | 21.7 | 0.7 | 0.52 | 24.5 | 22.8 | 19.9 |
| Srinivasan & Ebenezer | 32.2 | 30.4 | 28.6 | 0.84 | 0.74 | 26.1 | 24.2 | 23.1 |
| Ching, et al | 32.6 | 31.2 | 29.7 | 0.74 | - | - | - | - |
| Chan, et al | 31.8 | 30.8 | 29.7 | 0.86 | 0.755 | **28.1** | **26.7** | **25** |
| Fuzzy Cellular Automata | 31.8 | 30.5 | 29.2 | **0.88** | 0.757 | 26.5 | 25.3 | 24.1 |
| ITMF | 30.6 | 29.7 | 28.5 | 0.86 | 0.9422 | 23.7 | 22.9 | 21.9 |
| **STMDF-AD** | **32.7** | **31.3** | **30.8** | 0.87 | **0.9466** | 24.8 | 24.4 | 23.7 |

**Table 7** PSNR comparison of Lena images filtered with various filters from [38] and STMDF-AD

| Algorithm | ND (50%) | ND (60%) | ND (70%) | ND (80%) | ND (90%) |
|---|---|---|---|---|---|
| Srinivasan & Ebenezer | 26.8 | 26.1 | 25.5 | 24.49 | N/A |
| Chan, et al | 27.3 | 26.1 | 24.1 | 22.39 | 25.4 |
| Wang, et al | 28.5 | 27.3 | 26 | 24.53 | N/A |
| Fuzzy Cellular Automata | 28.7 | 27.6 | 26.1 | 24.68 | N/A |
| ITMF | 30.6 | 29.7 | 28.5 | 27.40 | 25.11 |
| **PA (STMDF-AD)** | **32.7** | **31.3** | **30.8** | **29.41** | **27.42** |

**Table 8** PSNR comparison of the top performing filters from the literature against the STMDF-AD

| ND (%) | Chan, et al [17] | LPC- SMF [40] | OCS [41] | EPA [42] | SAWM [43] | DAM [44] | CM [45] | FEMF [46] | ITMF [47] | (STMDF-AD) |
|---|---|---|---|---|---|---|---|---|---|---|
| 50 | N/A | **31.393 (Boat)** | 30.63(Lena) | 34.10(Lena) | **33.82(Lena)** | 32.78(Lena) | 33.85(Lena) | 33.28(Lena) | 30.6348 (Lena) | 30.3454 (Boat) 32.7293 (Lena) **24.8058 (Bridge)** |
| 60 | N/A | **N/A** | 30.55(Lena) | **32.66(Lena)** | 32.32 (Lena) | 31.24(Lena) | 32.11(Lena) | 31.64(Lena) | 29.7 (Lena) | 31.3494 (Lena) |
| 70 | 29.3 (Lena), **25.0 (Bridge)** | 28.133 (Lena), 26.775 (Boat) | 29.71(Lena) | **31.03(Lena)** | 30.69 (Lena) | 29.68(Lena) | 30.71(Lena) | 30.18(Lena) | 28.5 (Lena) | 30.7609 (Lena) **27.2360 (Boat)** 24.0122 (Bridge) |
| 80 | N/A | 25.836 (Lena) 24.555 (Boat) | 27.95(Lena) | 29.01(Lena) | 28.84 (Lena) | 27.95(Lena) | 28.59(Lena) | 28.47(Lena) | 27.40 (Lena) | **29.3444 (Lena)** **26.1193 (Boat)** |
| 90 | 25.4 (Lena) 21.5 (Bridge) | 24.316 (Lena) 22.220 (Boat) | 25.58(Lena) | 26.04(Lena) | 26.17(Lena) | 25.54(Lena) | 26.02(Lena) | 25.94(Lena) | 25.11 (Lena) | **27.4923 (Lena)** **24.3986 (Boat)** **21.7225 (Bridge)** |

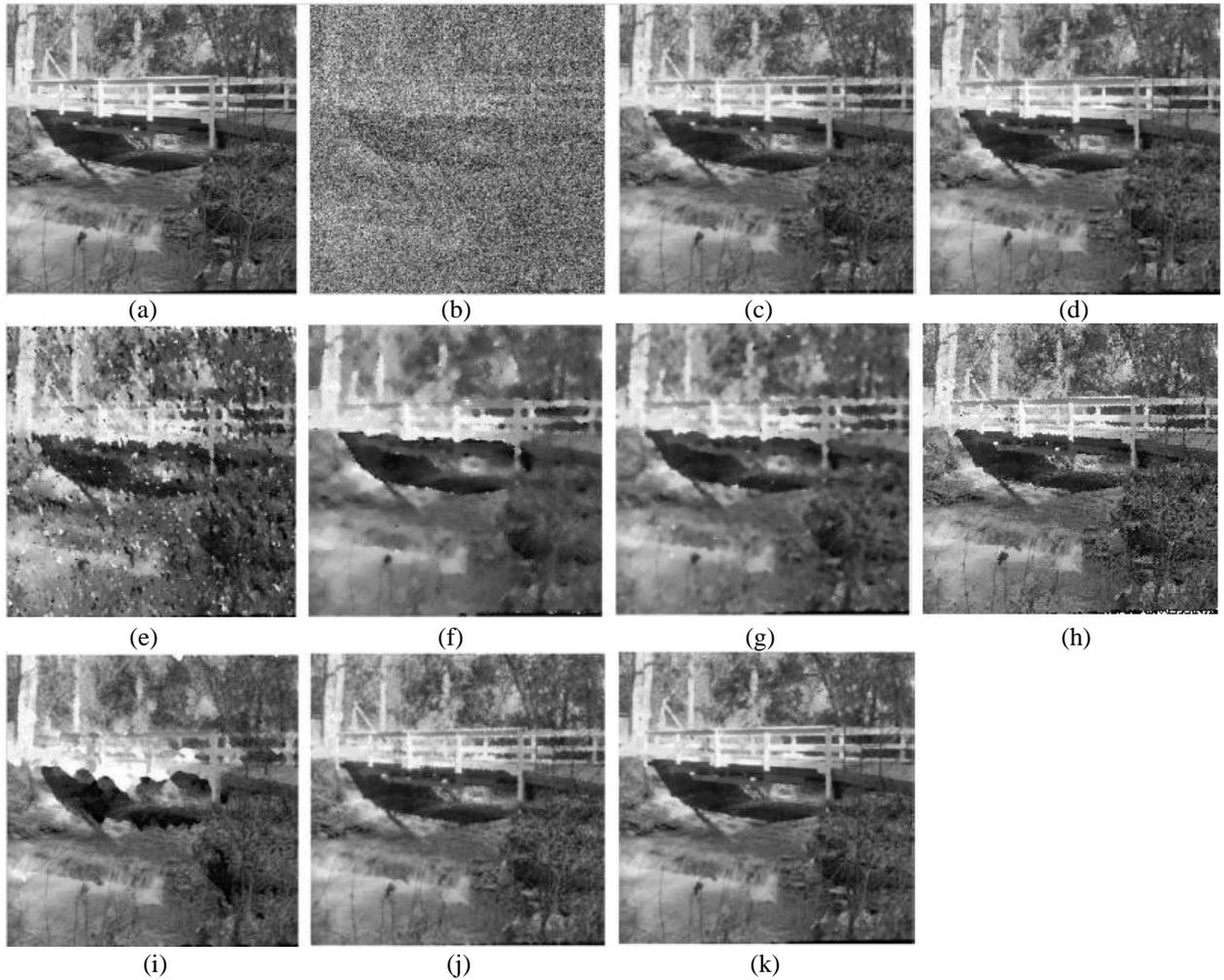

**Fig. 8** (a) Original Boat image (b) image corrupted with salt and pepper noise (ND = 70%) (c) filtered image using fuzzy cell automata (d) STMDF-AD (e) DDBSM (f) NASM (g) ISM (h) ITMF (i) PSM (j) Srinivasan (k) Chan et al

The edge preservation properties of the algorithm are appreciated at extreme levels of noise density when filtering the corrupted images. The failure of the other switching or decision-based filters is unavoidable as their limits become apparent at extreme high noise densities. The entropy guided threshold along with the energy minimization attributes of the proposed approach avoids successive smoothing in already treated regions with preserved edges, especially at high noise densities where differentiation between noise and signal becomes even more difficult to achieve in conventional schemes.

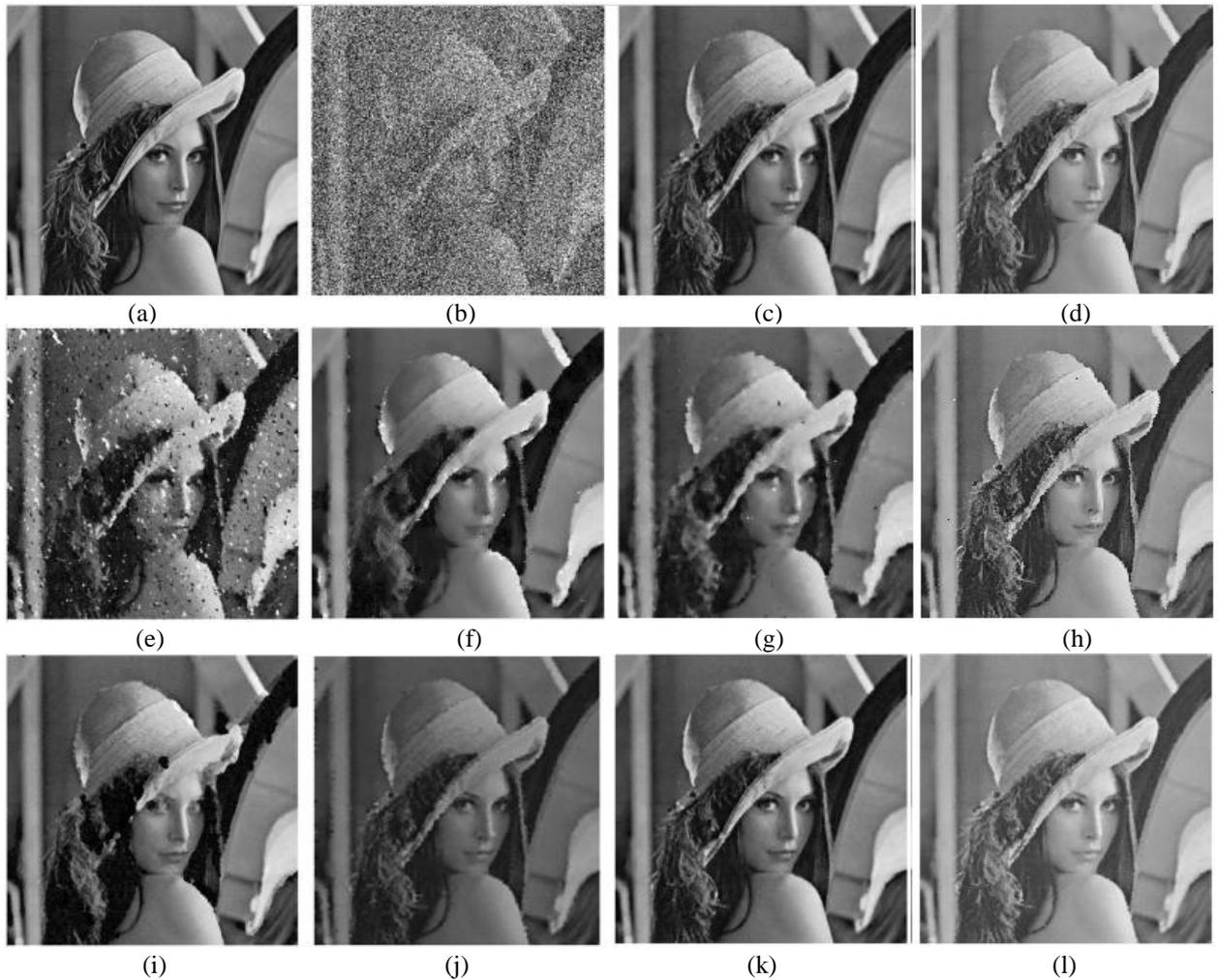

**Fig. 9** (a) Original Boat image (b) image corrupted with salt and pepper noise (ND = 70%) (c) filtered image using fuzzy cell automata (d) STMDF-AD (e) DDBSM (f) NASM (g) ISM (h) ITMF (i) PSM (j) Srinivasan (k) Chan et al (l) Ching, et al

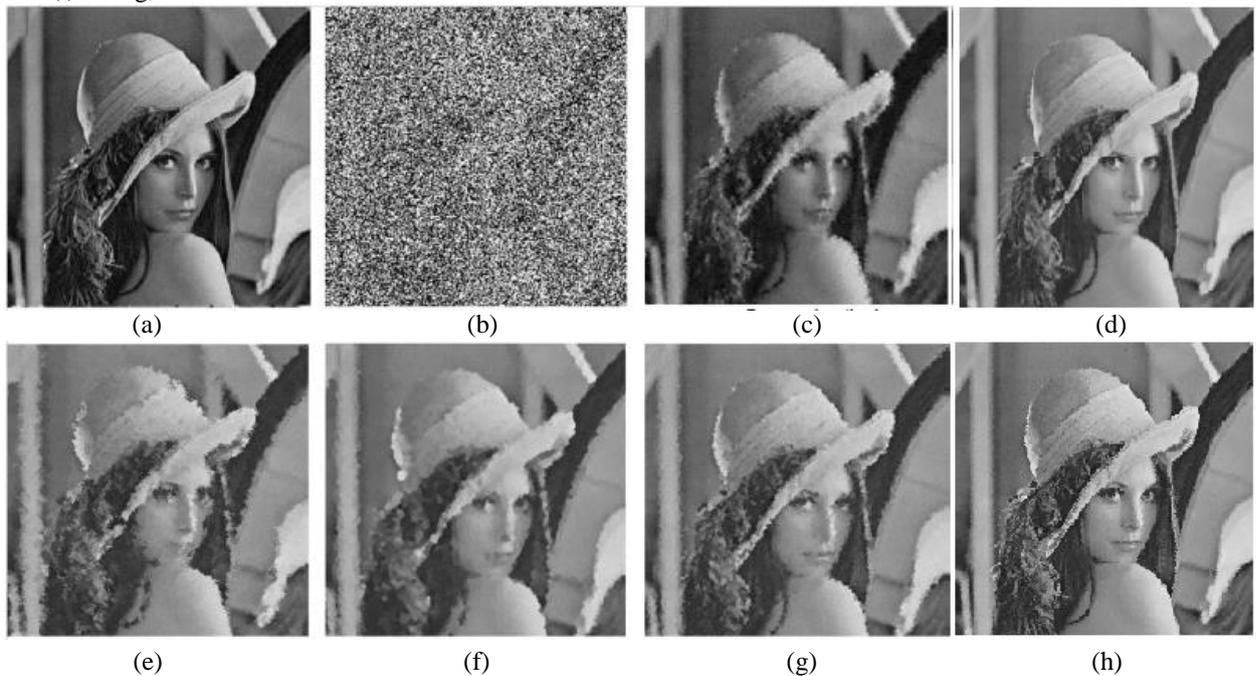

**Fig. 10** (a) Original Boat image (b) image corrupted with salt and pepper noise (ND = 80%) (c) filtered image using fuzzy cell automata (d) STMDF-AD (e) Chan, et al (f) Srinivasan (g) Wang, et al (h) ITMF

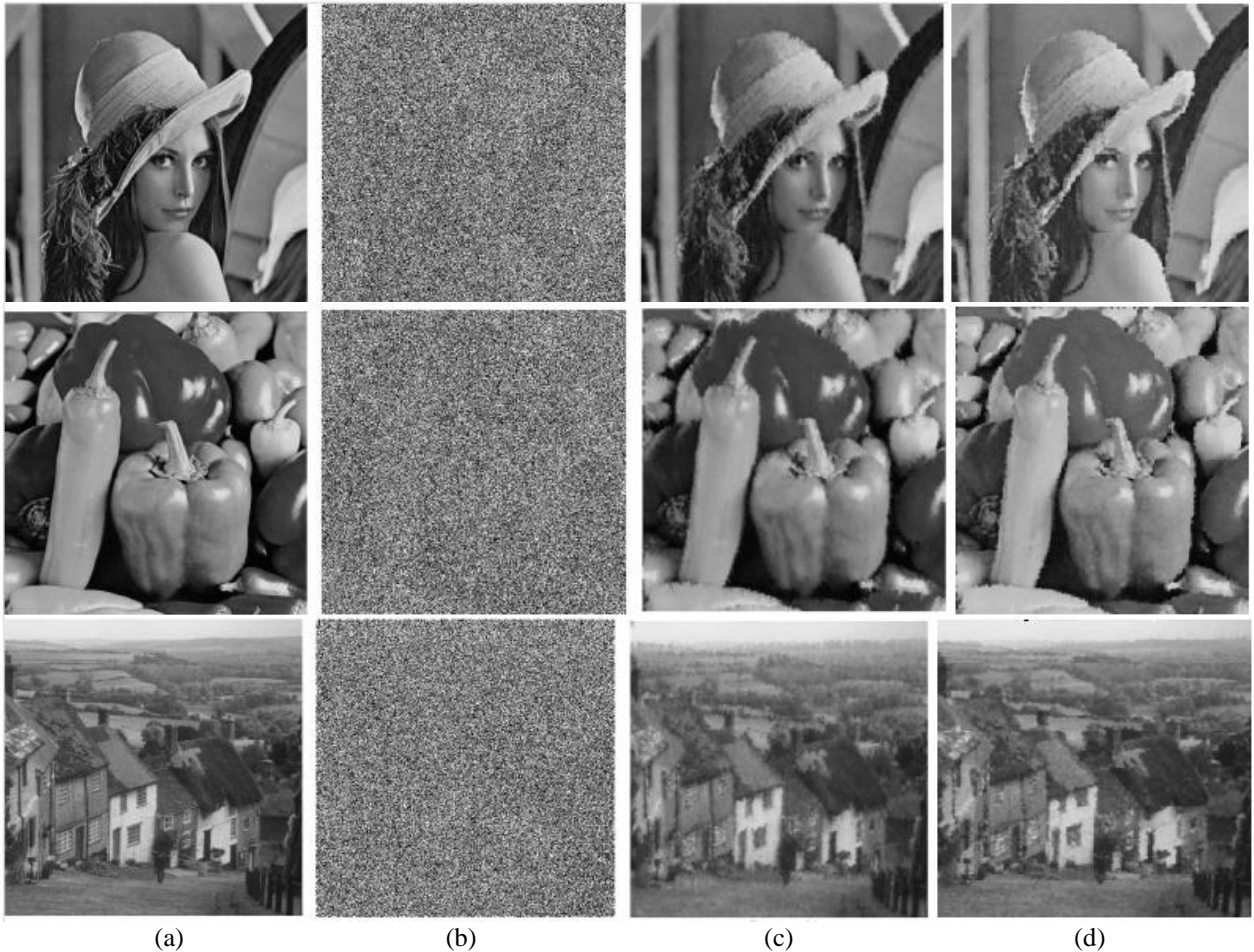

| (a) | (b) | (c) | (d) |

**Fig. 11** (a) Original Lena, Peppers & Gold Hill images (b) images corrupted with salt and pepper noise (ND = 90%) (c) Filtered images using fuzzy cell automata (d) STMDF-AD

## 4  Conclusion

This report has presented the theoretical considerations and detailed experimental results of the algorithm proposed in the paper titled, "Entropy-guided Switching Trimmed Mean-boosted Anisotropic Diffusion Filter [48]". Other variations of the approach are also mentioned and will be treated in much more detail in future works.